\documentclass[10pt,twocolumn,letterpaper]{article}

\usepackage[pagenumbers]{wacv} 

\usepackage{amsmath}
\usepackage{amssymb}
\usepackage{booktabs}
\usepackage{graphicx}
\usepackage{epstopdf}

\usepackage[pagebackref,breaklinks,colorlinks,]{hyperref}

\usepackage{booktabs}
\usepackage{paralist} 

\usepackage[accsupp]{axessibility}  
\usepackage{float}

\usepackage{adjustbox}

\usepackage{orcidlink}
\PassOptionsToPackage{table,usenames}{xcolor}
\usepackage{caption} 
\usepackage[framemethod=TikZ]{mdframed} 
\usepackage{helvet}
\usepackage[most]{tcolorbox} 
\usepackage{colortbl} 
\usepackage{multirow} 
\usepackage{enumitem}
\usepackage{fontawesome}
\usepackage{wrapfig}
\usepackage{mdframed}

\definecolor{Green}{rgb}{0.0, 0.5, 0.0}
\definecolor{Red}{rgb}{1.0, 0.0, 0.0} 
\definecolor{ForestGreen}{rgb}{0.13, 0.55, 0.13} 

\newcommand{\increase}[1]{(\textcolor{ForestGreen}{+#1})}
\newcommand{\increasenoparent}[1]{\textcolor{ForestGreen}{+#1}}

\newcommand{\decreasenoparent}[1]{\textcolor{red}{-#1}}

\usepackage{colortbl}

\definecolor{shadecolor}{gray}{0.9} 
\definecolor{examplegray}{gray}{0.95} 
\definecolor{lightgray}{gray}{0.9}
\definecolor{darkblue}{RGB}{31, 78, 121}   
\definecolor{darkgreen}{RGB}{78, 121, 31}  
\definecolor{lightgray}{RGB}{236,236,236}  
\definecolor{framecolor}{RGB}{100,100,100} 
\definecolor{darkgray}{RGB}{64,64,64}
\definecolor{darkred}{RGB}{139,0,0}
\definecolor{darkyellow}{RGB}{236,165,40}

\definecolor{darkpurple}{RGB}{75,0,130}
\definecolor{darkorange}{RGB}{255,140,0}

\definecolor{lightblue}{RGB}{173,216,230} 
\definecolor{skyblue}{RGB}{135,206,235}   
\definecolor{steelblue}{RGB}{70,130,180}  

\definecolor{apricot}{RGB}{255,224,178}   
\definecolor{orange}{RGB}{255,165,0}      
\definecolor{darkorange}{RGB}{255,140,0}  

\definecolor{lightgreen}{RGB}{144,238,144} 
\definecolor{olivedrab}{RGB}{107,142,35}  
\definecolor{forestgreen}{RGB}{34,139,34} 

\definecolor{lightgrey}{RGB}{211,211,211} 
\definecolor{darkgrey}{RGB}{169,169,169}  
\definecolor{dimgrey}{RGB}{105,105,105}   

\definecolor{lightpurple}{RGB}{230, 230, 250} 
\newcommand{\methodname}{\textcolor{dimgrey}{\textsc{VideoGameBunny}}\xspace}
\newcommand{\methodnameWhite}{\textcolor{white}{\textsc{VideoGameBunny}}\xspace}

\newcommand{\llavaonesix}{{{LLaVA-1.6-34b}}\xspace}

\definecolor{WatermarkPurple}{RGB}{150, 33, 247}
\definecolor{DoorGray}{RGB}{140, 126, 191}
\definecolor{MuralBlue}{RGB}{70, 130, 180}
\definecolor{SoldierRed}{RGB}{255, 28, 25}
\definecolor{InventoryGreen}{RGB}{108, 167, 89}
\definecolor{LightOrange}{RGB}{255, 154, 49}

\newcommand{\bunny}{Bunny\xspace}

\newcommand{\stwo}{S\textsuperscript{2}\xspace}

\newcommand{\geminiOneprovision}{Gemini-1.0-pro-vision\xspace}
\newcommand{\gptFourV}{GPT-4V\xspace}
\newcommand{\llamathree}{Llama-3\xspace}

\newcommand{\geminiOneFivePro}{Gemini-1.5-Pro\xspace}
\newcommand{\gptFourO}{GPT-4o\xspace}

\usepackage{enumitem}
\usepackage{fontawesome}
\usepackage{mdframed}
\usepackage{colortbl}

\usepackage{pifont}
\newcommand{\cmark}{\ding{51}}%
\newcommand{\xmark}{\ding{55}}%
\usepackage{colortbl}
\usepackage{soul}

\usepackage{mathptmx}
\definecolor{shadecolor}{gray}{0.9} 
\definecolor{examplegray}{gray}{0.95} 
\definecolor{lightgray}{gray}{0.9}
\definecolor{darkblue}{RGB}{31, 78, 121}   
\definecolor{darkgreen}{RGB}{78, 121, 31}  
\definecolor{lightgray}{RGB}{236,236,236}  
\definecolor{framecolor}{RGB}{100,100,100} 
\definecolor{darkgray}{RGB}{64,64,64}
\definecolor{darkred}{RGB}{139,0,0}
\definecolor{darkyellow}{RGB}{236,165,40}
\definecolor{darkerdarkorange}{rgb}{0.8, 0.4, 0.0}

\definecolor{lightblue}{RGB}{173,216,230} 
\definecolor{skyblue}{RGB}{135,206,235}   
\definecolor{steelblue}{RGB}{70,130,180}  

\definecolor{apricot}{RGB}{255,224,178}   
\definecolor{orange}{RGB}{255,165,0}      
\definecolor{darkorange}{RGB}{255,140,0}  

\definecolor{lightgreen}{RGB}{144,238,144} 
\definecolor{olivedrab}{RGB}{107,142,35}  
\definecolor{forestgreen}{RGB}{34,139,34} 

\definecolor{lightgrey}{RGB}{211,211,211} 
\definecolor{darkgrey}{RGB}{169,169,169}  
\definecolor{dimgrey}{RGB}{78,78,78}   

\definecolor{lightpurple}{RGB}{230, 230, 250} 

\usepackage{colortbl}
\usepackage{tcolorbox}

\usepackage[frozencache=true,cachedir=.]{minted} 
\tcbuselibrary{minted,breakable}

\newminted{json}{
    breaklines,
    breakanywhere,
    fontsize=\tiny,
    numbersep=5pt,
    linenos
}
\definecolor{keycolor}{RGB}{19,110,194}    
\definecolor{valuecolor}{RGB}{15,15,15}    
\definecolor{numcolor}{RGB}{176,0,64}      

\definecolor{lightgreyyy}{HTML}{EFEFEF}

\newcommand{\smallcolorbox}[2]{\raisebox{0.2em}{\smash{\colorbox{#1}{\textcolor{white}{\textbf{\scriptsize #2}}}}}}

\newtcolorbox{ebox}[2][]{
  colback=lightgray,
  colframe=darkgray,
  colbacktitle=darkgray,
  coltitle=white,
  title=#2,
  #1
}

\newmdenv[
  skipabove=\topsep,
  skipbelow=\topsep,
  middlelinewidth=1pt,
  roundcorner=5pt,
  frametitlebackgroundcolor=gray!20,
  backgroundcolor=gray!5
]{custommdframed}

\usepackage{pifont} 

\newcommand{\todo}[1]{{\textcolor{red}{#1}}}

\usepackage[capitalize]{cleveref}
\crefname{section}{Sec.}{Secs.}
\Crefname{section}{Section}{Sections}
\Crefname{table}{Table}{Tables}
\crefname{table}{Tab.}{Tabs.}

\def\wacvPaperID{182} 
\def\confName{WACV}
\def\confYear{2025}

\def\maketitlesupplementary
   {
   \onecolumn 
   \begingroup 
   \centering
   \Large
   \textbf{\maketitle}\\[0.5em]
   Supplementary Material \\[1.0em]
   \endgroup 
   }

\begin{document}
\title{\textbf{\methodname}: Towards vision assistants for video games}

\author{%
Mohammad Reza Taesiri \\
University of Alberta \\
\texttt{mtaesiri@gmail.com} \\
\and
Cor-Paul Bezemer \\
University of Alberta \\
\texttt{bezemer@ualberta.ca} \\
}

\maketitle

\begin{abstract}
Large multimodal models (LMMs) hold substantial promise across various domains, from personal assistance in daily tasks to sophisticated applications like medical diagnostics. 
However, their capabilities have limitations in the video game domain, such as challenges with scene understanding, hallucinations, and inaccurate descriptions of video game content, especially in open-source models.
This paper describes the development of \methodname, a LLaVA-style model based on \bunny, specifically tailored for understanding images from video games.
We release intermediate checkpoints, training logs, and an extensive dataset comprising 185,259 video game images from 413 titles, along with 389,565 image-instruction pairs that include image captions, question-answer pairs, and a JSON representation of 16 elements of 136,974 images. Our experiments show that our high quality game-related data has the potential to make a relatively small model outperform the much larger state-of-the-art model LLaVa-1.6-34b (which has more than 4x the number of parameters). Our study paves the way for future research in video game understanding on tasks such as playing, commentary, and debugging.
Code and data are available at:  
\url{https://videogamebunny.github.io/}
\end{abstract}

\begin{figure}[ht]
\begin{ebox}{\methodnameWhite understands game context}
\centering
\includegraphics[width=\textwidth,height=!]{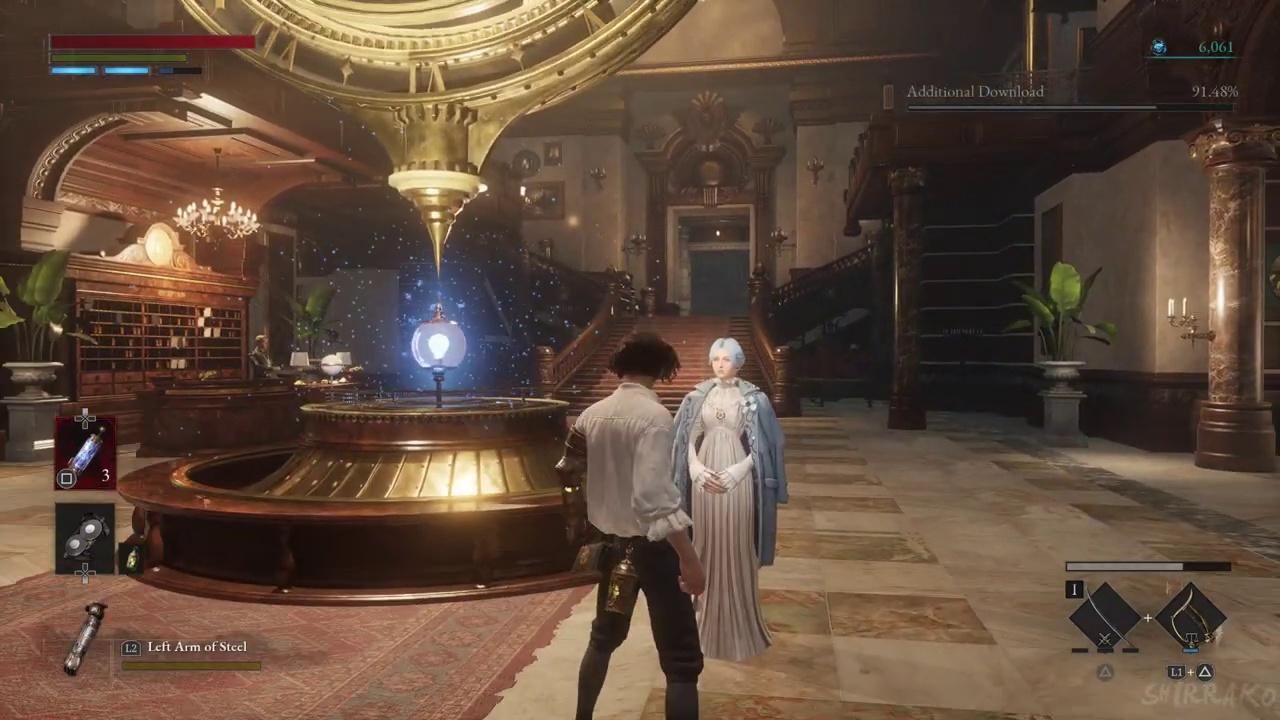}
\raggedright
\raggedright
\small
\textcolor{darkblue}{\textbf{Question:}} {\textbf{Are there any visible glitches or errors in the game environment?}} \\
{\fontsize{7.8pt}{11pt}\selectfont\textbf{\methodname:} \textbf{\textcolor{blue}{(D)}}: No, there are no apparent glitches. \textcolor{Green}{\cmark} \\}
{\fontsize{7.7pt}{11pt}\selectfont\textbf{\textcolor{darkerdarkorange}{\bunny}:}\textbf{\textcolor{blue}{(B)}}: Yes, the glowing orb is clipping through the counter.  \textcolor{Red}{\xmark} \\}
\textbf{\textcolor{darkpurple}{\llavaonesix:}} \textbf{\textcolor{blue}{(C)}} Yes, the `Additional Download' progress bar seems stuck. \textcolor{Red}{\xmark}
\end{ebox}
\caption{\methodname is a model specifically fine-tuned on video game content, enabling it to understand game contexts and respond to related questions more accurately.}
\label{fig:teaser_1}
\end{figure}
    
\section{Introduction}

The video game industry is projected to be valued at \$321 billion by 2026~\cite{wef2023gaming, grandview2023gaming} and continues to contribute more to the global economy.
Generative Artificial Intelligence (GenAI)~\cite{jo2023promise} is rapidly spreading across various sectors, disrupting the ways in which many traditional tasks are executed~\cite{bommasani2021opportunities, eloundou2023gpts}. In the realm of gaming, GenAI has the potential to enhance many aspects, such as  providing a better in-game experience by generating more realistic conversations with non-playable characters (NPCs)~\cite{NVIDIA_Newsroom_2023} or  better graphical assets~\cite{poole2022dreamfusion, Holson_2023, zhang2024clay, chen2024meshanything}.

Large language models (LLMs) and large multimodal models (LMMs) represent advancements in GenAI with the potential to function as vision assistants and solve complex problems across various domains~\cite{achiam2023gpt, team2023gemini, anthropic2024claude3}. In video games, LMMs can offer significant benefits for both in-game experiences and game development.
In-game, LMMs can serve as vision assistants, enhancing players’ experiences by guiding them through tasks like crafting new items~\cite{mashable2024copilot}. They also have the potential to narrate the game, summarize events, and highlight critical gameplay moments~\cite{reddit_gpt4o_prototype}.
For game development, LMMs have the potential to assist in detecting bugs~\cite{taesiri2024glitchbench}, creating bug reports, and deploying automated in-game bots that interact with the environment~\cite{tan2024towards}.
These applications require robust models capable of understanding game content.

Despite advances and promises, existing LMMs, particularly open-source models, encounter challenges in accurately understanding game content, such as scenes and world physics~\cite{taesiri2024glitchbench} (e.g., see ~\cref{fig:teaser_1}). 

In this study, we make the first important step towards addressing these challenges by releasing a suite of datasets specifically designed for video game content and introducing \methodname, a model trained for video game content understanding. Our study centers on the following research questions:

\begin{compactenum}
    \item[] \textbf{(RQ1)} \textit{Which type of data has the potential to improve the model's performance?}
    \item[] \textbf{(RQ2)} \textit{Which data type mixture strategy improves the model's performance the most?}
    \item[] \textbf{(RQ3)}  \textit{How does \methodname perform compared to state-of-the-art (SOTA) open-source models on game understanding tasks?}
\end{compactenum}

Our main contributions are as follows:

\begin{compactenum}
    \item We release \methodname, a model specifically fine-tuned for video game question-answering tasks.
    \item We release a suite of datasets containing 185,259 video game images from 413 games, featuring various gameplay elements and graphical styles. Our datasets include 389,565 image-instruction pairs with captions, question-answering tasks, and JSON representations of images (see \cref{sec:data}).
    \item We conduct experiments to demonstrate the effectiveness of different instruction datasets and their impact on the model's performance (see \cref{sec:results}).
    \item We release a replication package containing the training logs and intermediate checkpoints at \url{https://videogamebunny.github.io/}.
\end{compactenum}

\section{Background and Related Work}

\subsection{Large multimodal models}

Large multimodal models (LMMs) enhance large language models (LLMs) by incorporating additional modalities such as images or audio, enabling them to process multimodal inputs and generate textual outputs. The role of the language model is to comprehend user instructions and produce responses based on the additional modality inputs provided.
Standard approaches to create LMMs involve combining pre-trained models with different modalities via projection layers. These layers can be implemented using simple mechanisms such as multilayer perceptrons (MLP)~\cite{li2024llava, liu2023improvedllava} or transformer layers~\cite{li2023blip}. Alternatively, a resampler module like Perceiver~\cite{jaegle2021perceiver, alayrac2022flamingo, laurenccon2024obelics} or  Qformer~\cite{dai2024instructblip, zhu2023minigpt} selectively chooses features to reduce the number of visual tokens based on the context and instruction, enhancing efficiency and maintaining performance.

In this study, we focus on LMMs that accept input images and text to produce responses, particularly using the LLaVA-style architecture~\cite{liu2023llava}, which is one of the most popular methods~\cite{liu2024llavanext, lin2023video, he2024efficient, mckinzie2024mm1}. This architecture employs an MLP layer to integrate vision tokens with a language model.

\subsection{Instruction following data}

Large models trained on massive corpora of text, such as GPT-3~\cite{brown2020language}, T5~\cite{raffel2020exploring}, and PaLM~\cite{chowdhery2023palm}, are not inherently instruction-following, meaning they do not respond to user queries. To enable these models to follow user instructions and answer queries, they usually undergo a process called instruction tuning~\cite{ouyang2022training, zhang2023instruction}. This process involves fine-tuning the models to handle specific user instructions, such as questions or commands, allowing them to respond appropriately based on the given instructions.

In the multimodal context, particularly for models that accept visual inputs, there are various types of visual instruction-following data, such as detailed descriptions, conversational style question answering (Q\&A), and complex reasoning.
Researchers have explored diverse approaches to generate such data, including the use of academic text-oriented visual Q\&A datasets~\cite{dai2024instructblip}.
The LLaVA model~\cite{li2024llava} demonstrated that leveraging a strong text-only LLM and an image dataset annotated with object names and bounding box information can be converted into effective visual instruction-following data.

\subsection{LLMs and LMMs in video games}

LLMs have shown strong promise for integration with games for a wide range of tasks, from content creation to game-playing agents~\cite{xu2024survey, hu2024survey, gallotta2024large, wu2024spring,zhang-lu-2024-adarefiner,ma2023large,shao2024swarmbrain, taesiri2022large, wang2023voyager, wang2023describe, wang2023jarvis, zheng2023steve}. Large multimodal models (LMMs) can further enhance this integration by providing richer context inputs such as images and videos to enable broader applications.
Projects like Cradle~\cite{tan2024towards}, which focuses on playing Red Dead Redemption 2 with GPT-4V~\cite{achiam2023gpt} showcase LMMs' abilities to identify objects, characters, and environmental features and assist in controlling the game. 
Beyond gameplay, LMMs have found applications in game testing~\cite{taesiri2024glitchbench, taesiri2022large}, where they are leveraged for detecting and interpreting video game bugs.

Our study is the first to explore enhancing an LMM's general game understanding, rather than focusing on a specific game or task. We use screenshots from 413  games, aiming to improve capabilities across various game-related tasks by developing broader game comprehension skills.

\subsection{Empirical analysis of large multimodal models}

Some previous studies have conducted experiments to see how different architectural components or data sources affect the general performance of large multimodal models~\cite{tong2024cambrian, mckinzie2024mm1, li2024llavanext-ablations, laurenccon2024matters}. For example, McKinzie et al.\cite{mckinzie2024mm1} found that the input resolution of the input image plays a crucial role in improving performance, and Laurençon et al.\cite{laurenccon2024matters} found that utilizing cross-attention between image and language is more effective than the adapter-based method.

We are the first to systematically investigate the impact of different instruction-following datasets and their combinations on the performance of LMMs in game understanding tasks.

\section{\methodname Model Architecture}

In this section, we describe the architectural choices and configurations behind our model, \methodname. \methodname is based on \bunny~\cite{he2024efficient}, a family of efficient and high-performing LLMs known for their competitive or superior performance on various benchmarks compared to many open-source alternatives.

\bunny follows the same principle as LLaVA~\cite{li2024llava, liu2023improvedllava} for the integration of image inputs. Using a shallow network of multilayer perceptrons (MLPs) as the projection layer, vision embeddings extracted from a strong pre-trained vision model are processed and provided as image tokens for the language model. This technique effectively leverages pre-trained vision and language models, allowing them to work together efficiently.

\begin{figure}[t]
  \centering
  \includegraphics[width=\columnwidth]{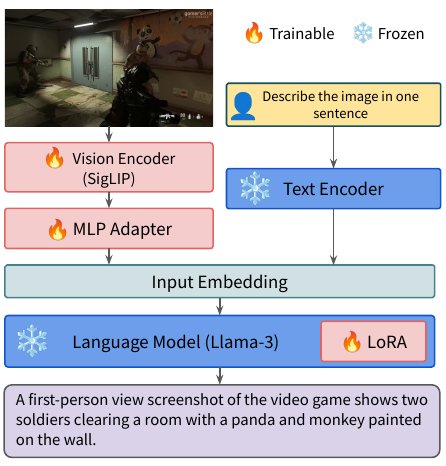}
  \caption{Architecture overview of \methodname. An image input and a textual instruction are fed into the language model to produce a response. The image is passed through a separate pre-trained vision encoder and a projection layer to align the embedding space between the two models. \raisebox{-0.2ex}{\includegraphics[scale=0.02]{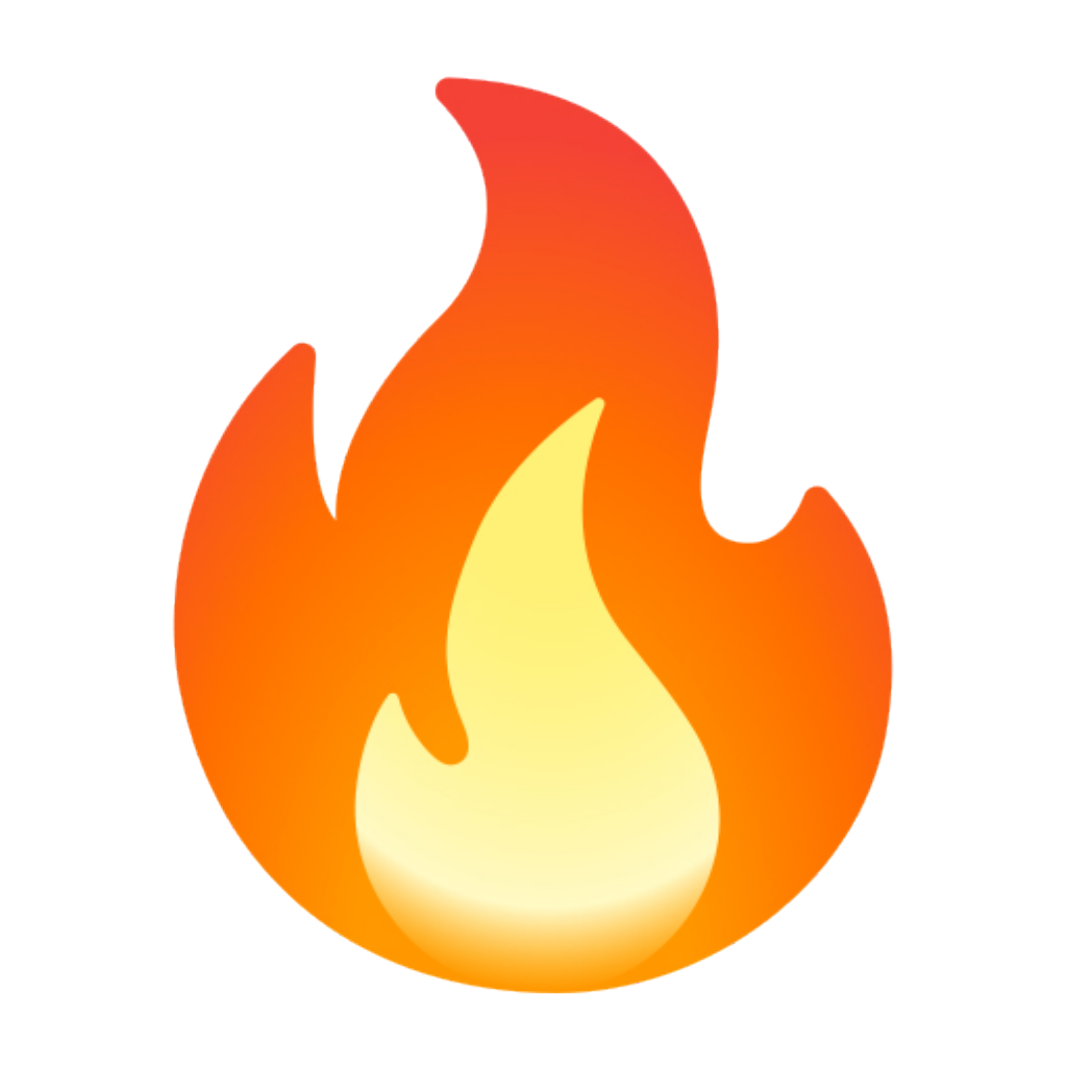}}  and \raisebox{-0.2ex}{\includegraphics[scale=0.02]{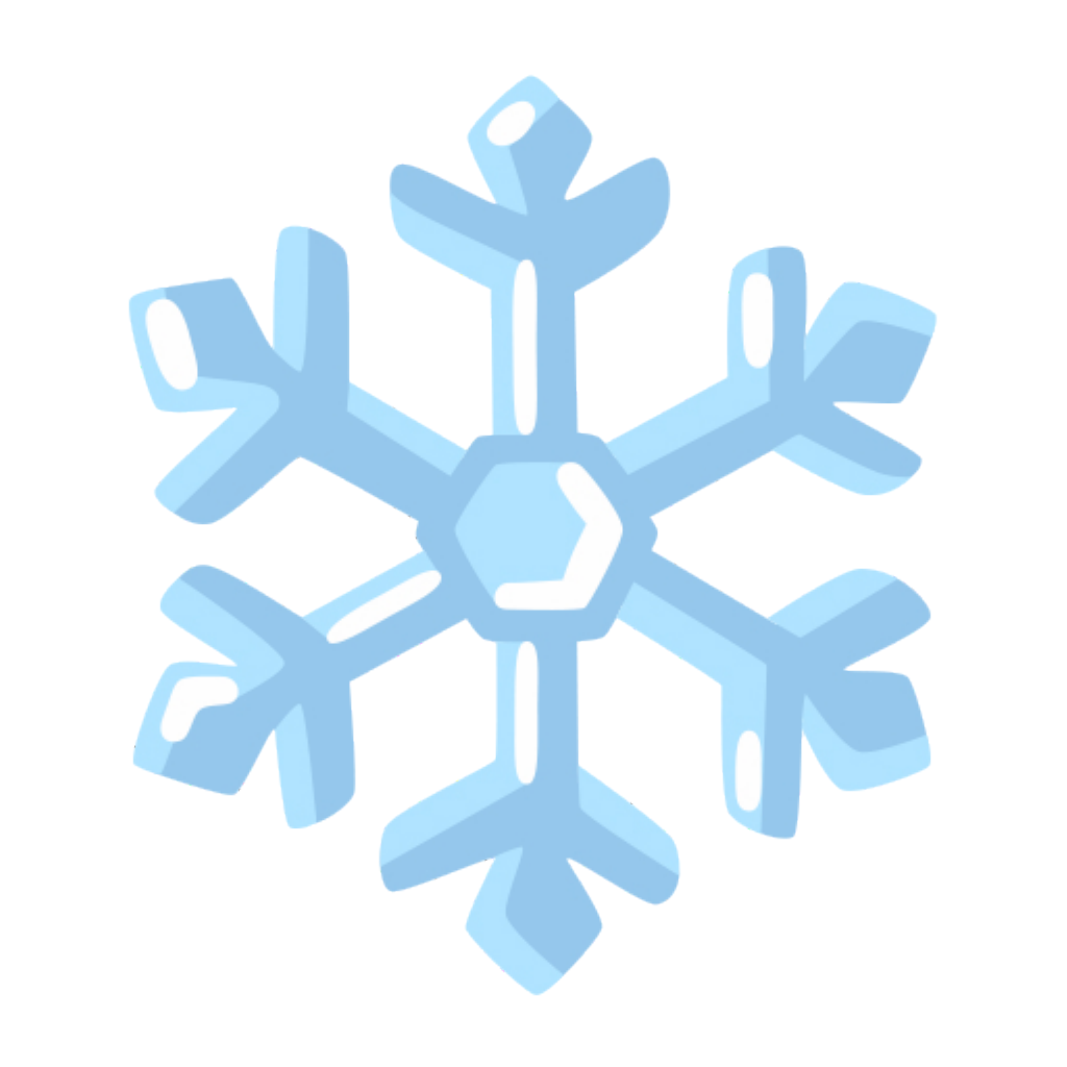}} icons show trainable and frozen layers respectively}
  \label{fig:bunny_arch}
\end{figure}

\bunny offers various combinations of vision and language models and supports images with resolutions up to $1152 \times 1152$ pixels.
For creating \methodname, we selected \bunny configurations that deliver the best performance~\cite{he2024efficient} while being small enough to run on a consumer-grade graphics card. We use LLama-3-8B~\cite{llama3modelcard} as the language model and SigLIP~\cite{zhai2023sigmoid} with the \stwo wrapper~\cite{shi2024we} for the vision encoder.
The \stwo wrapper extracts features from an input image at various scales to form a multi-scale feature.
This is potentially useful since video games often contain visual elements at different scales, from tiny UI icons to large objects. A multi-scale feature could  capture these diverse elements.
\cref{fig:bunny_arch} shows the architecture of \methodname.

\section{Instruction-following Data for Video Game Content}
\label{sec:data}

\begin{figure*}[t!]
\centering
\begin{subfigure}{0.245\textwidth}
  \includegraphics[width=\linewidth, trim=0 25 0 25, clip]{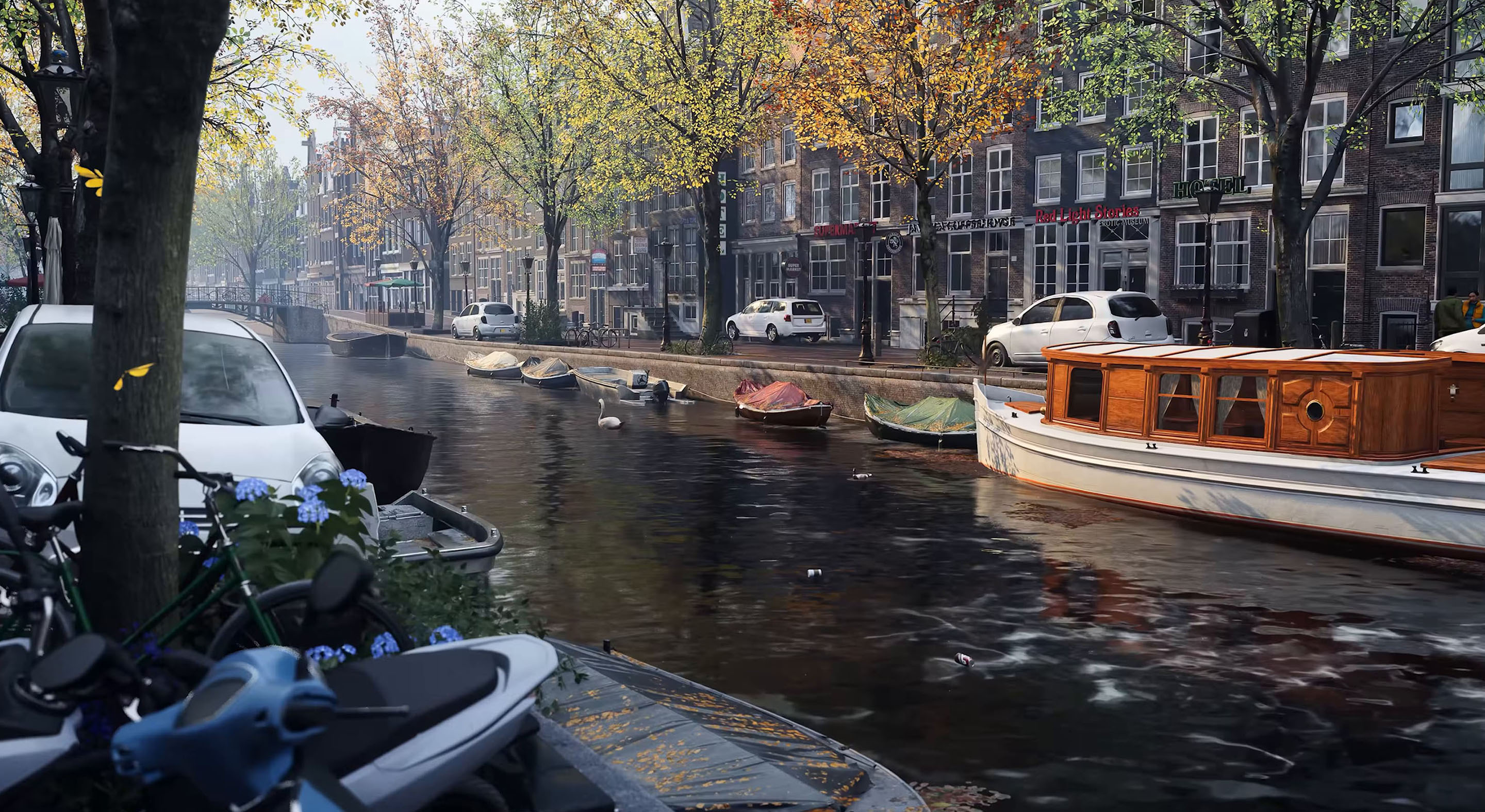}
\end{subfigure}
\hfill 
\begin{subfigure}{0.245\textwidth}
  \includegraphics[width=\linewidth, trim=0 25 0 25, clip]{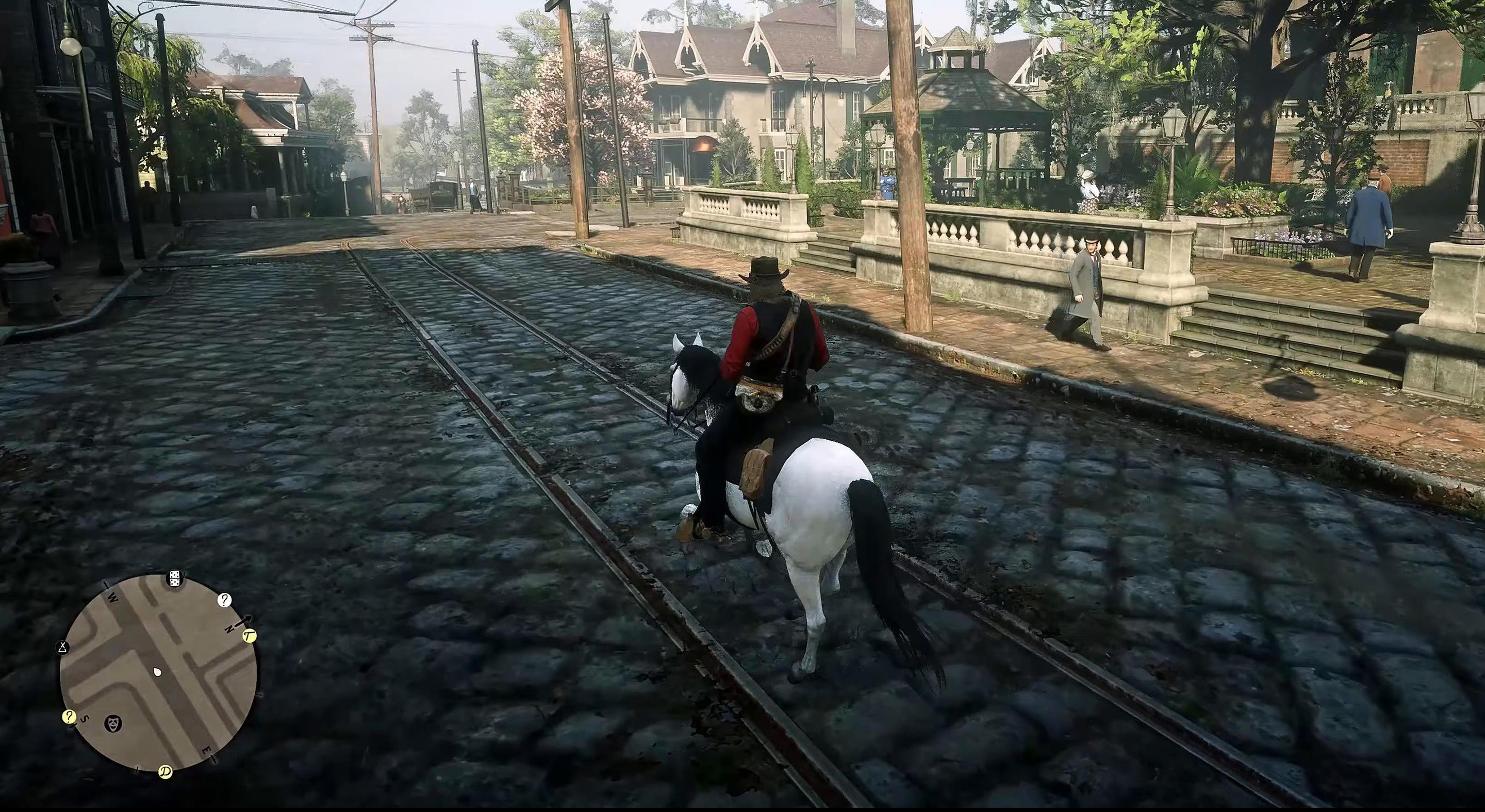}
\end{subfigure}
\hfill
\begin{subfigure}{0.245\textwidth}
  \includegraphics[width=\linewidth, trim=0 25 0 25, clip]{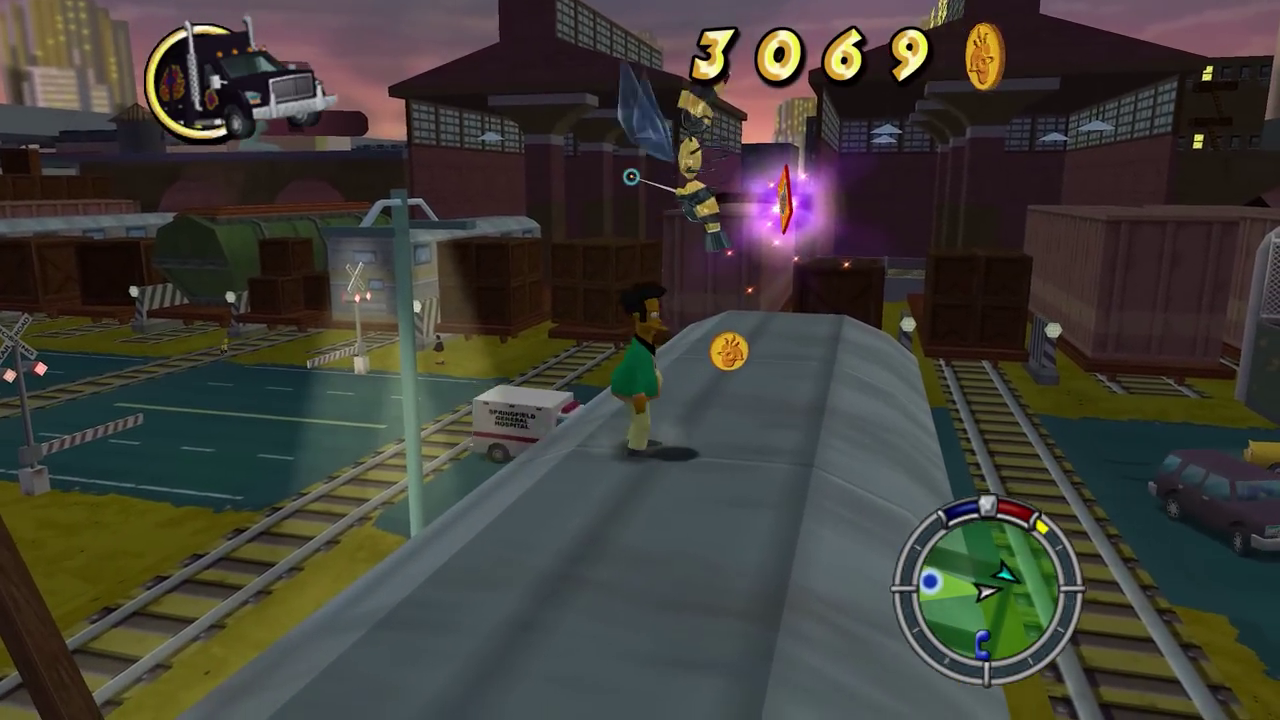}
\end{subfigure}
\hfill
\begin{subfigure}{0.245\textwidth}
  \includegraphics[width=\linewidth, trim=0 25 0 25, clip]{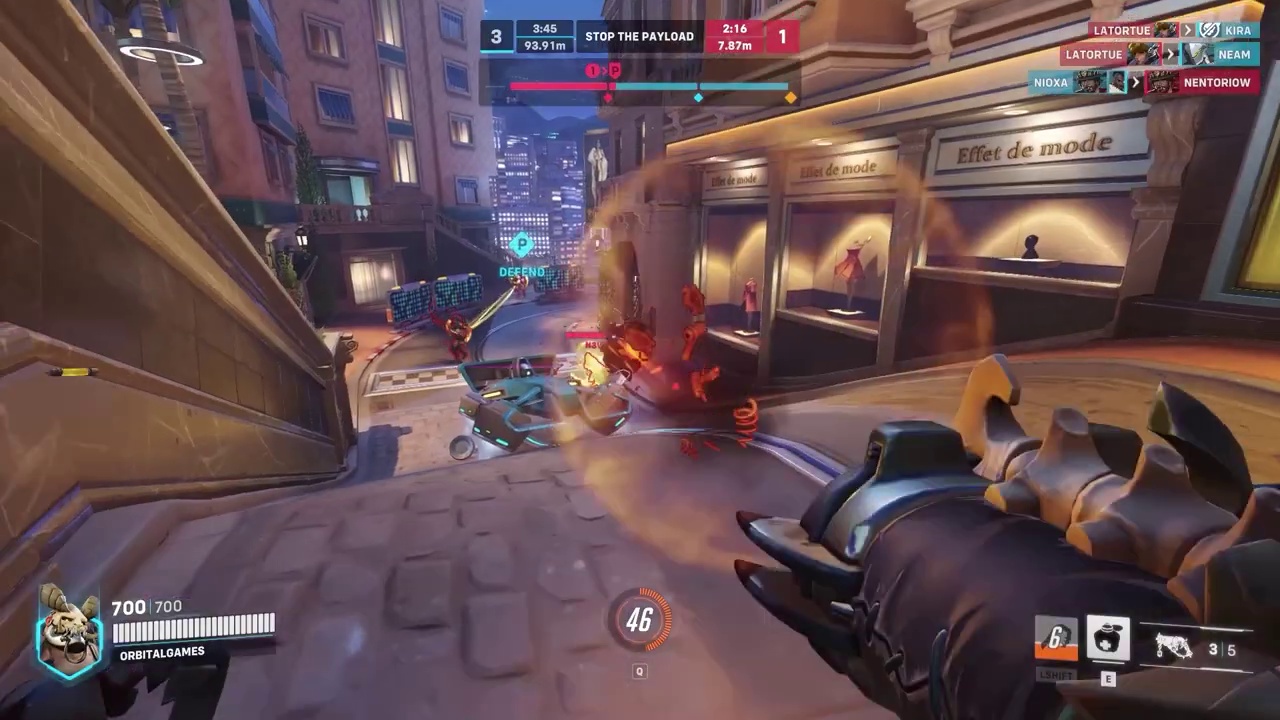}
\end{subfigure}
\caption{Our dataset includes sample video game images that showcase a wide range of characters, environments, mechanics, camera viewpoints, and artistic styles. These styles vary from western to contemporary and futuristic, and from realistic to fantasy settings.}
\label{fig:sample_images}
\end{figure*}

\begin{figure}
    \centering
    \includegraphics[width=\linewidth]{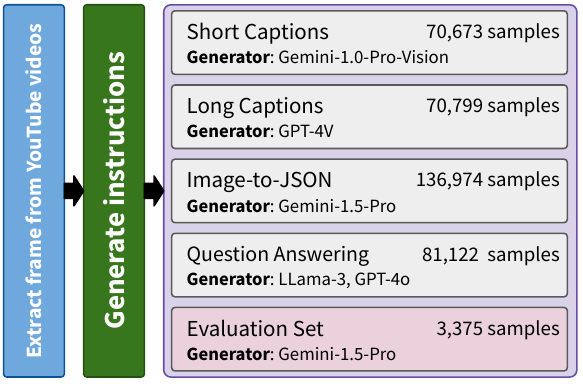}
    \caption{Overview of the dataset generation process.}    
    \label{fig:data_collection_and_generation}
\end{figure}

One of the main challenges limiting the ability of open-source models to generalize effectively to video game content is the lack of instruction-following data specific to video games in public datasets. Our goal is to collect game-specific data to address this challenge.
In this section, we explain the process of collecting and generating game-specific instruction-following data.

\subsection{Video game images}

We collect images from YouTube by searching for \emph{gameplay walkthroughs} with \emph{Full-HD}, \emph{4K}, and \emph{8K} quality. These high-resolution videos ensure that downsampled frames retain more information and details compared to lower quality videos. We randomly sample frames from the downloaded videos and label them with the corresponding game name.
In total, our dataset contains 185,259 images from 413 different video games, encompassing various genres, graphic styles, and gameplay mechanics. \cref{fig:sample_images} shows some sample images from our dataset, and \cref{suppfig:histo_games} shows the distribution of images per game.

\subsection{Generating instructions}

Following previous studies~\cite{liu2023llava, liu2023improvedllava, xiao2024florence}, we employ another robust model to generate instructions in the form of user queries and responses for images in our dataset. We categorize the instructions into four types: \textit{short captions} (\textit{70,673} samples), \textit{long captions} (\textit{70,799} samples), \textit{image-to-JSON} (\textit{136,974} samples) and \textit{image-based question answering} (\textit{81,122} samples). In this section, we explain how we generate each type of instructions. 
\cref{fig:data_collection_and_generation} shows an overview of the data generation process.

\subsubsection{Image captioning}

Image captioning is a basic form of instruction-following that generates a description of the input image.
An image caption can be short and concise, providing a high-level overview of the image, or very detailed, covering fine-grain details. Our dataset includes both forms of image captioning to meet user queries, whether they seek a detailed caption or a short summary. In addition, it includes a structured version where the image is described in 16 predefined fields. 

\textbf{Short captions}:
We use \geminiOneprovision to obtain short descriptions of a subset of images in our dataset, which includes 70,673 images. We use the ``\textit{Describe the image}’’ prompt, which generates captions typically consisting of one or a few sentences.

\textbf{Long captions}:
While short captions provide a high-level overview of the image, it lacks many details in the image which might be useful for the user. To address this, we use  \gptFourV  to get detailed captions of all images in the previous section (see \cref{prompt:gpr4v_long_caption} for the used prompt).

\textbf{Image-to-JSON}:
Another comprehensive method for describing images is converting them into a JSON format.
This approach summarizes an input image into a JSON structure, with each key describing an element from the image, such as \textit{characters in the image} or \textit{description of Game UI}.
Unlike typical captioning, this method provides a template that must be filled. If the image lacks a certain element, that part remains empty, indicating the absence of that element in the image. This ensures a more detailed and structured representation of the image content.

Another benefit of describing an image in JSON is that this structured representation facilitates integration with other systems. 
JSON is a widely adopted format for sharing information between different software systems. Summarizing the image as JSON can help integrate LMMs in other systems, such as software testing pipelines, potentially verifying the game output and ensuring that the visual output has the desired properties and information.

\begin{figure*}[t]
    \centering
    \includegraphics[width=\textwidth]{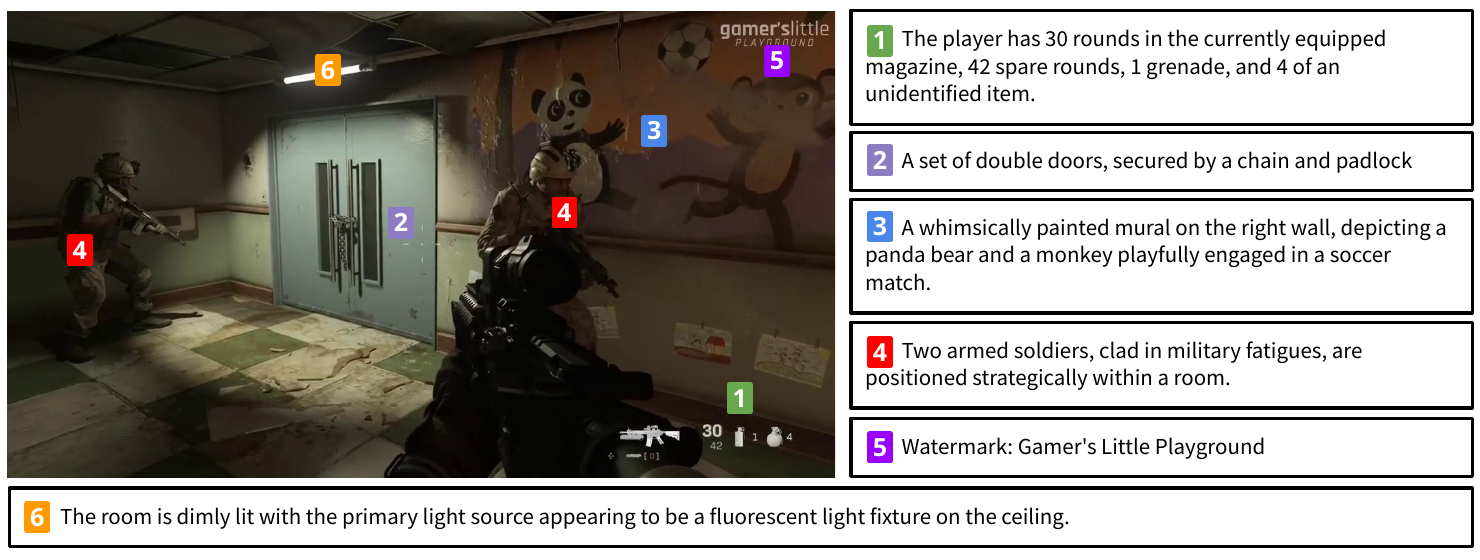}
    \caption{Sample information extracted for the image-to-JSON dataset by \geminiOneFivePro. Each sample contains detailed information ranging from minor details to high-level descriptions, such as:
    \smallcolorbox{InventoryGreen}{1} player inventory,
    \smallcolorbox{DoorGray}{2} \smallcolorbox{MuralBlue}{3} details about the environment,
    \smallcolorbox{SoldierRed}{4} non-player characters,
    \smallcolorbox{WatermarkPurple}{5} the screenshot's watermark, and
    \smallcolorbox{LightOrange}{6} lighting.}
    \label{fig:dataset_overview_Sample}
\end{figure*}

To create the image-to-JSON dataset, we use \geminiOneFivePro with instructions (\cref{prompt:image_to_json_gemini}) to convert a given image into a JSON file with hierarchical levels of detail and information.
The JSON file contains 16 elements that capture both high-level and fine-grained details of the image. 
These keys are chosen to capture game-specific elements from the image in isolation, which can be used for downstream applications, such as game testing.
It starts with an overall summary of the image and then moves to specific aspects such as detailed character descriptions (including facial expressions and clothing), weather information, summaries of UI and player inventory, objects in the scene, and lighting and environmental effects. \cref{supptab:json_keys} shows the keys included in the JSON output. 
Our dataset contains 136,974 Image-JSON pairs. \cref{fig:dataset_overview_Sample} shows a sample of information extracted from an image.

\subsubsection{Question-Answering conversations}

Moving beyond simple image descriptions, a general and capable model should be able to respond to user questions based on the content of the image. Below, we describe how we generate such data for each image (1) from its long caption and (2) directly from the image itself. 

\textbf{\llamathree-assisted visual instruction data generation}:
We use \llamathree to convert long captions generated by \gptFourV into a series of question-answering conversations. This approach is similar to the original LLaVA~\cite{liu2023llava} method, but instead of using an object's name and its bounding box information, we directly utilize long captions. 

While long captions provide a rich source of information, they lack the structure of question-answer formats.
For example, if a caption describes a person in the image with specific details, such as clothing, an LLM can generate a question like, \textit{``What is the color of the dress of the person in the image?''}
By utilizing a strong text-only model, we can transform each caption into a multi-turn conversation between a user and an assistant.

We use \llamathree-70B to transform \gptFourV captions into question-answer conversations, with the prompt shown in \cref{prompt:llama3}.
The prompt requires questions to directly relate to the image description. We create 496,469 question-answer pairs for 70,232 images, grouping questions for each image into a multi-turn conversation.

\textbf{Image-based question-answering}:
We use \gptFourO to generate questions and their answers based on an input image in a \emph{single} prompt.
In the prompt (\cref{prompt:gpt4o}), we first ask \gptFourO to examine the image and provide a detailed description of its content, then to generate relevant questions based on the content of the image and provide answers for each question.  In the prompt, we emphasize that the questions should focus on understanding the image to avoid questions that might not be directly relevant to the image.

\subsection{Evaluation dataset}

To assess model performance on video game understanding tasks, we created a multiple-choice question evaluation set using \geminiOneFivePro~\cite{team2023gemini}. This approach allows for an efficient comparison of various models. While \geminiOneFivePro offers significant advantages over open-source models for data generation, it does have limitations. We reduce noise in the generated questions as follows:

\begin{compactenum}
    \item \textbf{Initial Generation:} We use \geminiOneFivePro to create 4,000 questions across 10 categories related to video game content understanding (see \cref{tab:val_set_categories}).

    \item \textbf{Quality Assessment:}
    \begin{compactenum}
        \item Self-evaluation: We  test \geminiOneFivePro on its own questions and found it achieves an accuracy of 84\%.
        \item Manual validation: A random sampling of questions and answers revealed a 14\% error rate (incorrect or indeterminate answers).
    \end{compactenum}

    \item \textbf{Noise Reduction:}
    \begin{compactenum}
        \item We remove 625 samples that \geminiOneFivePro had answered incorrectly.
        \item We conduct a second manual analysis and found that the error rate dropped to 9\%.
    \end{compactenum}
\end{compactenum}

\begin{table*}[t]
    \centering
    \caption{Categories of questions in our dataset, along with a sample question for each category.}
    \resizebox{0.9\textwidth}{!}{
    \begin{tabular}{llr}
    \textbf{Category} & \textbf{Description} & \textbf{Count} \\
    \midrule
    \rowcolor[HTML]{FFFFFF} \textbf{Action Understanding} & Recognizing and describing the actions taking place within the image. & 356 \\
    \rowcolor[HTML]{FFFFFF} & \textit{Sample: What action is the character in the foreground performing?} & \\ 
    \rowcolor[HTML]{EFEFEF} \textbf{Anomalies and Glitches} & Identifying errors, bugs, glitches, or placeholder elements within the game environment. & 223 \\
    \rowcolor[HTML]{EFEFEF} & \textit{Sample: Describe any anomalies or glitches present in the image.} & \\ 
    \rowcolor[HTML]{FFFFFF} \textbf{Character Analysis} & Recognizing characters, understanding their roles, and interpreting their expressions and poses. & 312 \\
    \rowcolor[HTML]{FFFFFF} & \textit{Sample: What is Aloy's emotional state based on her facial expression?} & \\ 
    \rowcolor[HTML]{EFEFEF} \textbf{Common Sense Reasoning} & Understanding the image using general knowledge and everyday logic. & 430 \\
    \rowcolor[HTML]{EFEFEF} & \textit{Sample: Based on the score and time remaining, which team is likely to win the match?} & \\ 
    \rowcolor[HTML]{FFFFFF} \textbf{Gameplay Mechanics} & Understanding the rules and mechanics that govern the game. & 273 \\
    \rowcolor[HTML]{FFFFFF} & \textit{Sample: What game mechanic is most likely being utilized by the player character?} & \\ 
    \rowcolor[HTML]{EFEFEF} \textbf{OCR and UI} & Reading and interpreting on-screen text and user interface elements. & 334 \\
    \rowcolor[HTML]{EFEFEF} & \textit{Sample: What is written in the caption box at the bottom of the image?} & \\ 
    \rowcolor[HTML]{FFFFFF} \textbf{Miscellaneous} & Any other type of question that does not fit into the previous categories. & 239 \\
    \rowcolor[HTML]{FFFFFF} & \textit{Sample: What material are the containers in the image primarily made of?} & \\ 
    \rowcolor[HTML]{EFEFEF} \textbf{Scene Understanding} & Recognizing and interpreting the overall environment or setting in the image. & 566 \\
    \rowcolor[HTML]{EFEFEF} & \textit{Sample: The racetrack depicted in the image is set in what type of environment?} & \\ 
    \rowcolor[HTML]{FFFFFF} \textbf{Small Details} & Identifying and interpreting small but significant details within the image. & 356 \\
    \rowcolor[HTML]{FFFFFF} & \textit{Sample: What color is the jacket worn by the character in the foreground?} & \\ 
    \rowcolor[HTML]{EFEFEF} \textbf{Spatial Reasoning} & Testing the ability to understand spatial relationships of objects present in the image. & 286 \\
    \rowcolor[HTML]{EFEFEF} & \textit{Sample: What is the spatial relationship between the two red markers visible in the image?} & \\  
    \bottomrule
\end{tabular}
    }
    \label{tab:val_set_categories}
\end{table*}


\section{Experiments}
\label{sec:experiments}

In this section, we describe our experiments to explore how our collected instruction-following datasets can improve a model's understanding of game context. We focus on three research questions:

\textbf{(RQ1) \textit{Which type of data has the potential to improve the model's performance?}}
In addressing this question, we fine-tune \bunny using a single dataset at a time to observe overall performance trends. Since the primary goal of this experiment is to identify general trends, we fine-tune \bunny on different subset sizes for each dataset only \emph{once}. We increase the subset size from 2K to 60K samples and stop the experiment if we observe a sharp decline in performance.

\textbf{(RQ2) \textit{Which data type mixture strategy improves the model's performance the most?}}
We evaluate different data mixing strategies at various sizes to see how both mixture and subset size change the performance of the model. We use the following four strategies:

\begin{compactenum}
    \item \textbf{Random:} We randomly sample without replacement from the combined dataset pool. This serves as a control group, using no specific selection strategy.
    \item \textbf{Equal:} We select an equal number of samples from each dataset to ensure a balanced representation.
    \item \textbf{Stratified:} Datasets are mixed based on \emph{video games},, maintaining the game distribution in the final dataset. This balances game representation and ensures diverse image types. We focus on game variety rather than instruction types. Games with insufficient samples are excluded.
    \item \textbf{Weighted:} We use the three most effective datasets from \textbf{RQ1}: image-based question-answering (\gptFourO), long captions, and image-to-JSON. We assign weights: 30\% each for \gptFourO and long captions, 40\% for image-to-JSON. This prioritizes valuable datasets to assess their impact on model performance.
\end{compactenum}

We fine-tune \bunny on the above dataset mixture strategies with sizes ranging from 2K to 30K. We repeat each experiment three times, using different samples for each strategy to report the mean performance and standard deviation. We stop at 30K since our smallest dataset (generated by \gptFourO) contains 10K samples, and at 30K, we will exhaust the \emph{Equal} and \emph{Weighted} strategies.

\textbf{(RQ3) \textit{How does \methodname perform compared to SOTA open-source models on game understanding tasks?}}
Building on insights from our experiments, we create \methodname, a model fine-tuned on a dataset of 50K image-instruction samples compiled from all previously introduced datasets.
To assess the effectiveness of fine-tuning a smaller model on game-specific data, we evaluate \methodname against \llavaonesix, a SOTA open-source model with 4.2$\times$ more parameters.

\textit{Experiment setup:}
We instruction tune \bunny  with LoRA~\cite{hu2021lora} using the PEFT~\cite{peft} library.
Given that \bunny has been trained on real images, we unfreeze the vision encoder (SigLIP~\cite{zhai2023sigmoid}) to adapt to the diverse visual styles of different games.
To prevent overfitting and memorization, we fine-tune for only one epoch in all experiments.

Given the importance of reproducibility and accessibility for all researchers, we perform all experiments on a \textbf{single NVIDIA A100 GPU (80GB)}, ensuring a balance between computational power and accessibility. The total GPU hours needed to conduct all experiments, including some preliminary tests, is approximately 900 hours, which is roughly \$2,000 when using cloud providers.

\section{Results}
\label{sec:results}

\begin{figure}[t]
  \centering
  \includegraphics[width=\columnwidth]{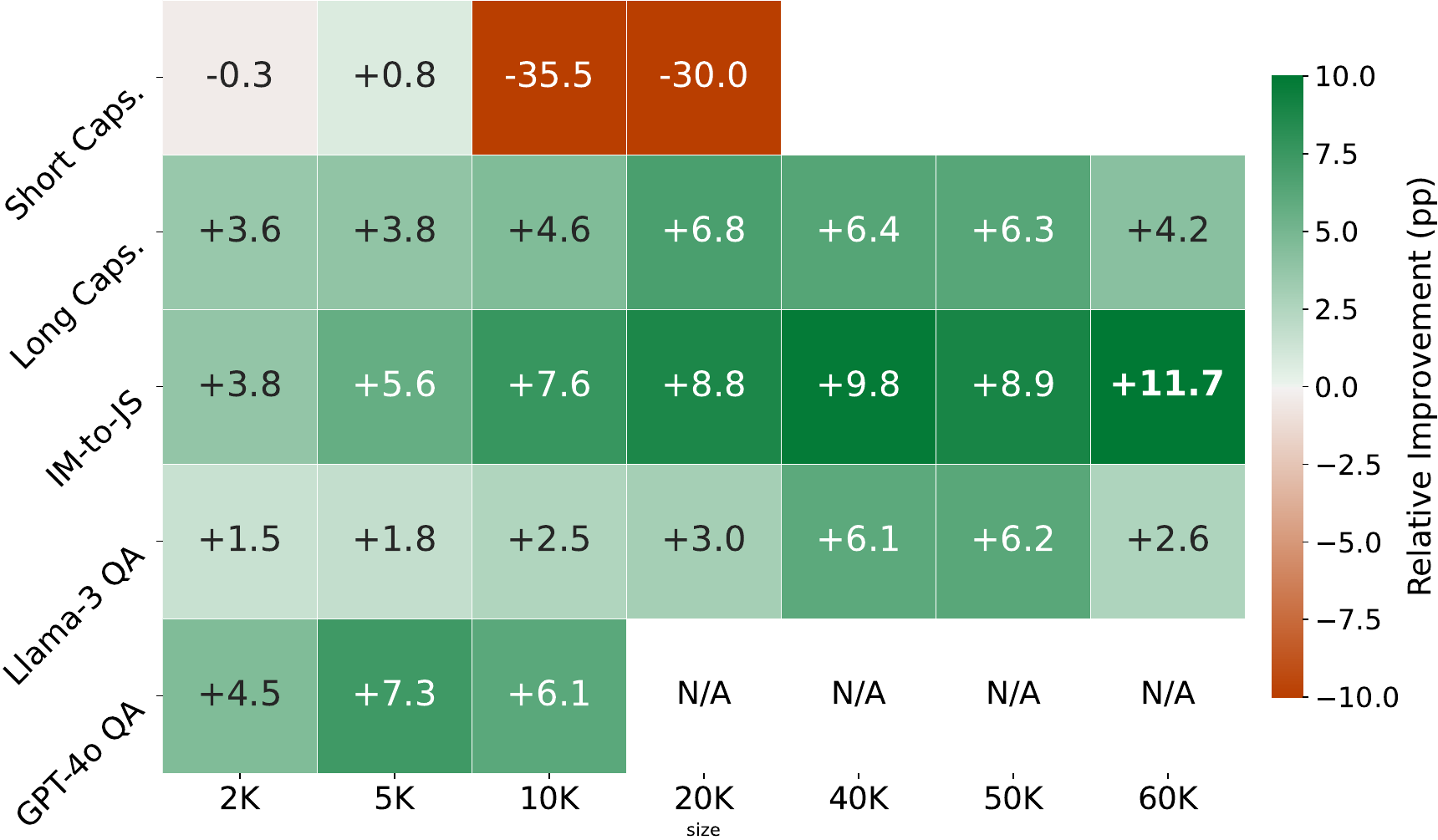}
\caption{Relative performance improvement (pp) of \bunny fine-tuned on different subsets of each dataset. The image-to-JSON dataset shows a strong positive trend, while the short captions dataset degrades performance. The best performance achieved in the experiment is highlighted in bold.}
  \label{fig:trends_in_data}
\end{figure}


\subsection*{RQ1:  Which type of data has the potential to improve the model's performance?}
\label{sec:trends_in_data}

\textbf{The image-to-JSON dataset has the greatest potential to improve the base model's performance.} \cref{fig:trends_in_data} shows the performance after fine-tuning \bunny using a single dataset at a time.
Fine-tuning on a subset of the image-to-JSON dataset shows the greatest improvements, as this leads to an accuracy above 82\% (\increasenoparent{8.7} percentage points (pp)  above the baseline of 73.3\%) for subset sizes over 10K, with the best performance achieved at 60K (\increasenoparent{11.7} pp).

\textbf{While all datasets lead to performance improvement, short captions can degrade it.} Fine-tuning \bunny on a dataset of 10K or 20K short captions degrades performance (\decreasenoparent{35.5} pp and \decreasenoparent{30} pp), suggesting that short captions do not contain enough signal for the models to improve and can negatively affect the model.

\subsection*{RQ2: Which data type mixture strategy improves the model's performance the most?}
\label{sec:mixture_strategies_results}

\begin{table}[t]
\centering
\caption{Performance of models fine-tuned on a mixture of data with various strategies. The Weighted strategy leads to better performance with smaller dataset sizes, but as size increases, all strategies perform similarly. We use a strategy similar to Weighted to train \methodname with 50K samples.}
\resizebox{\columnwidth}{!}{
    \begin{tabular}{lrrrr}
    \toprule
    \cmidrule(r){2-5}
    Size & \multicolumn{1}{c}{Random} & \multicolumn{1}{c}{Equal} & \multicolumn{1}{c}{Stratified} & \multicolumn{1}{c}{Weighted} \\
    \cmidrule(r){2-5}
    2K   & 76.7 ± 0.9 & 77.8 ± 0.8 & 78.0 ± 0.2 & 79.0 ± 0.6 \\
    5K   & 79.2 ± 0.4 & 79.9 ± 0.4 & 80.0 ± 0.5 & 79.8 ± 0.6 \\
    10K  & 79.8 ± 0.8 & 80.8 ± 0.6 & 80.8 ± 0.1 & 81.4 ± 0.5 \\
    20K  & 81.5 ± 0.1 & 81.3 ± 0.7 & 81.8 ± 0.8 & 82.3 ± 0.9 \\
    30K  & 81.8 ± 0.4 & 81.2 ± 1.1 & 81.6 ± 0.7 & 82.6 ± 0.3 \\
    \midrule
    50K  & -- & -- & -- & \textbf{85.1} \\
    \bottomrule
    \end{tabular}
}
\label{tab:mixture_data_results}
\end{table}


\textbf{There is a general improvement trend as we increase the size across all strategies.} \cref{tab:mixture_data_results} shows the performance of the models that were fine-tuned using our data mixture strategies. 
As we increase the dataset size, the mean performance of all mixtures improves. 
For instance, the \emph{Random} strategy improves from 76.7\%  at 2K samples to 81.9\% at 30K samples and the \emph{Weighted} strategy shows an improvement from 79.0\%  at 2K samples to 82.6\% at 30K samples.
This trend demonstrates the value of additional data regardless of the mixing strategy employed.


\textbf{As the size of dataset increases, different strategies perform similarly.}
The performance difference between various strategies converges as we increase the size of the datasets, and they perform similarly in terms of mean and standard deviations.
Yet, the \emph{Weighted} method achieves the highest average among other strategies (82.6\%).
This convergence suggests that the choice of mixing strategy becomes less critical as more data becomes available.
In contrast, smaller dataset sizes such as 2k indicate that the \emph{Weighted} strategy outperforms other mixture strategies, achieving an accuracy of 79.0 $\pm$ 0.6.

\textbf{Having a uniform distribution of games does not significantly improve performance.}
The \emph{Stratified} strategy, which aims to balance the representation of different games in the dataset, does not significantly enhance performance compared to other strategies.
For example, in the 2k dataset, the \emph{Stratified} strategy (78.0 $\pm$ 0.2) is outperformed by the \emph{Weighted} strategy (79.0 $\pm$ 0.6). Similarly, in the 30k dataset, the performance of both strategies is comparable (81.6 $\pm$ 0.7 vs 82.6 $\pm$ 0.3).

\textbf{Fine-tuning improves performance across all categories, with \emph{Anomalies and Glitches} improving the most.}
\cref{suppfig:results_category} shows that fine-tuning improves \bunny's performance across all  categories for almost all dataset sizes. The \emph{Anomalies and Glitches} and \emph{HUD and UI} categories improve the most, with average improvements of \increasenoparent{32.0} and \increasenoparent{21.0}, using a dataset size of 30K (\cref{tab:avg_per_size}).

\begin{table}[t]
\centering
\small
\caption{Average improvement for different sizes for each category}
\label{tab:avg_per_size}
\begin{tabular}{lrrrrr}
\toprule
Category/\textbf{Dataset Size}  & \textbf{2K} & \textbf{5K} & \textbf{10K} & \textbf{20K} & \textbf{30K} \\
\midrule
Action Understanding & 1.6 & 2.5 & 2.5 & 3.7 & 3.9 \\
Anomalies and Glitches & 23.4 & 33.0 & 33.2 & 34.0 & 32.0 \\
Character Analysis & 2.6 & 3.9 & 4.2 & 4.7 & 4.4 \\
Common Sense Reasoning & 3.7 & 4.2 & 3.8 & 4.3 & 4.0 \\
Gameplay Mechanics & 4.2 & 5.0 & 6.4 & 8.2 & 8.9 \\
HUD and UI & 9.3 & 12.9 & 16.5 & 18.9 & 21.0 \\
Miscellaneous & 7.2 & 7.9 & 9.6 & 9.9 & 9.8 \\
Scene Understanding & \decreasenoparent{-0.2} & 0.6 & 1.3 & 2.0 & 2.0 \\
Small Details & 0.3 & 1.2 & 2.4 & 3.4 & 3.0 \\
Spatial Reasoning & 5.3 & 6.2 & 7.1 & 7.8 & 7.4 \\
\bottomrule
\end{tabular}
\end{table}

\subsection*{RQ3: How does \methodname perform compared to SOTA open-source models on game understanding tasks?}
\label{sec:method_vs_sota}
\begin{figure}[t]
\begin{ebox}{Models comparison in video game understanding}
\centering
\includegraphics[width=\textwidth]{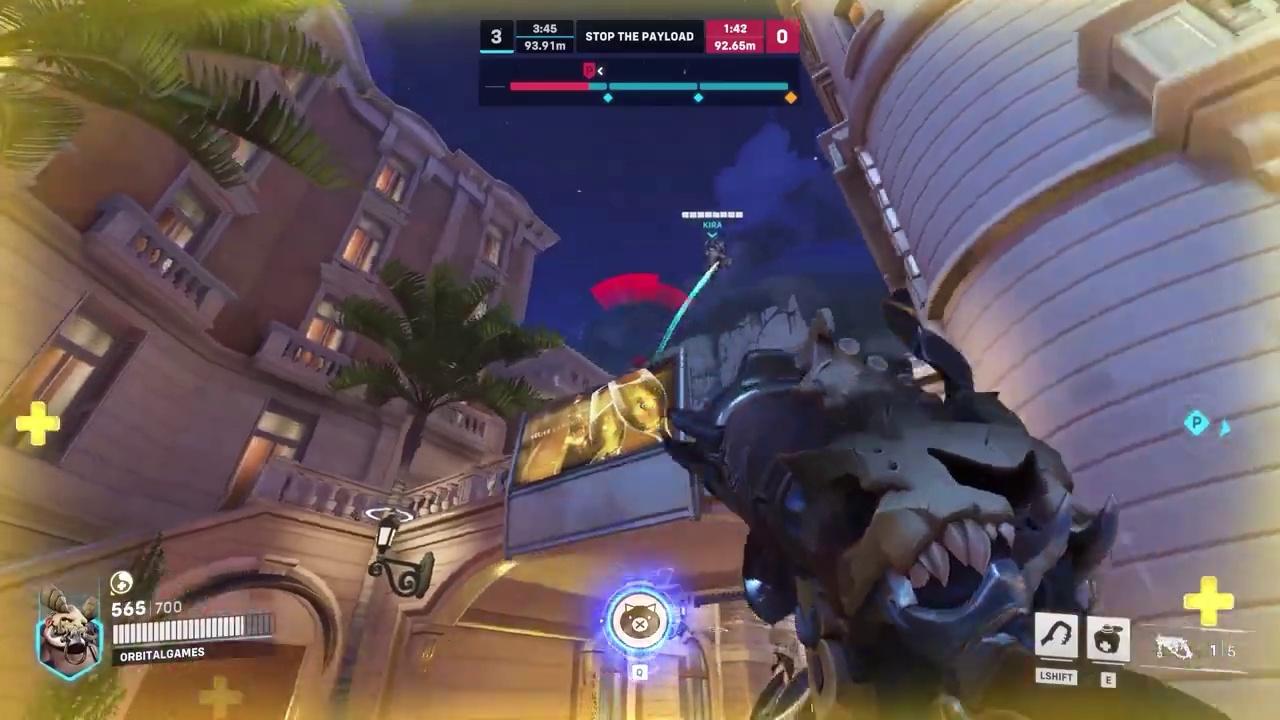}
\raggedright
\raggedright
\small
\textcolor{darkblue}{\textbf{Question:}} {\textbf{Based on the score and time remaining, which team is likely to win the match?}}  \\
{\fontsize{8.2pt}{11pt}\selectfont\textbf{\methodname:} \textbf{\textcolor{blue}{(B)}}: The blue team is likely to win \textcolor{Green}{\cmark} \\}
\textbf{\textcolor{darkerdarkorange}{\bunny}:} \textbf{\textcolor{blue}{(C)}}: The red team is likely to win. \textcolor{Red}{\xmark} \\
\textbf{\textcolor{darkpurple}{\llavaonesix:}} \textbf{\textcolor{blue}{(C)}}: The red team is likely to win. \textcolor{Red}{\xmark}
\end{ebox}
\caption{\methodname correctly utilizes information on the HUD to answer the question, while the basemodel \bunny and the larger model \llavaonesix fail to provide a correct answer.}
\label{fig:qual_sample_1}
\end{figure}

\methodname  achieves \textbf{85.1\%} (\cref{tab:mixture_data_results})  on the evaluation set, outperforming all trained models and surpassing various open-source models (\cref{tab:eval_results}). 
 It outperforms even \llavaonesix, despite its larger parameter count, by \increasenoparent{1.2}.
Breakdown of accuracy per category reveals that the most significant benefits come from game-specific categories, such as anomaly and glitch detection \increase{16.6} and HUD and UI \increase{3.0} (\cref{apptab:vgb_vs_llava}).
\cref{fig:qual_sample_1} shows a sample from evaluation set where \methodname correctly answers a common-sense reasoning question while both \bunny and \llavaonesix fail.

\begin{figure}[ht]
\begin{ebox}{Models comparison in image captioning}
\centering
\includegraphics[width=\textwidth]{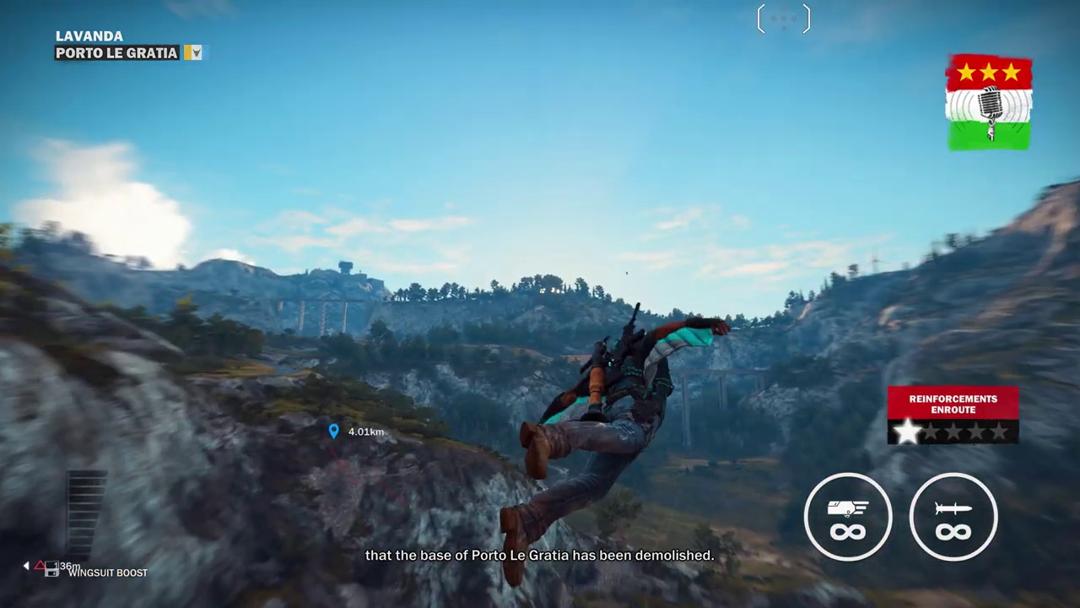}
\raggedright
\textbf{\textcolor{darkblue}{Question:}} \textbf{Write a short summary of the image}  \\
\textbf{\methodname:} The image is a screenshot from a video game, showing a character in mid-air, equipped with a wingsuit, flying over a mountainous landscape. The game's HUD elements are visible, including a distance marker and a weapon indicator. \ldots \\ 
\textbf{\textcolor{darkerdarkorange}{\bunny:}}  A video game scene shows a man in a \textcolor{red}{\emph{green shirt}} and blue pants jumping off a cliff. He is \textcolor{red}{\emph{holding a parachute}} and is in mid-air. \ldots \\
\textbf{\textcolor{darkpurple}{\llavaonesix:}} The image appears to be a screenshot from a video game, showing \textcolor{red}{\emph{a first-person perspective}} of a character in mid-air, likely after jumping or falling from a height. The character is \textcolor{red}{\emph{equipped with a parachute, which is open}} and slowing their descent. \ldots
\end{ebox}
\caption{\methodname provides a detailed description of video game content in the image, while other models show inaccuracies (highlighted in \textcolor{red}{\emph{red}}). Responses are truncated to save space.}
\label{fig:qual_sample_2}
\end{figure}

We also conduct an informal qualitative analysis, prompting the models with more open-ended questions, which underlines that \methodname demonstrates an improved understanding of video game content and provides better answers (\cref{fig:qual_sample_2}, and \cref{app:bugsbunny_qual_samples}).

\begin{table}[t]
\caption{Performance of various models on the evaluation set (\%). }
\label{tab:eval_results}
\centering
\resizebox{\columnwidth}{!}{%
    \begin{tabular}{lcrlcr}
        \toprule
        \textbf{Model} & \textbf{Accuracy} & & \textbf{Model} & \textbf{Accuracy} \\
        \midrule
        \text{Bunny-1.1-Llama-3-8B} & \multicolumn{1}{r}{73.3} & & \text{LLaVA-v1.5-13b} & \multicolumn{1}{r}{64.6} \\
        \text{\methodname} & \multicolumn{1}{r}{\textbf{85.1}} & & \text{LLaVA-v1.6-vicuna-13b} & \multicolumn{1}{r}{71.7} \\
       \text{LLaVA-v1.5-7b}  & \multicolumn{1}{r}{61.3} & & \text{LLaVA-v1.6-34b} & \multicolumn{1}{r}{83.9} \\
        \bottomrule
    \end{tabular}%
}
\end{table}

\section{Discussion}

\textbf{Potential negative impacts:}
Our study initiates the development of a model for understanding game content, with applications in game playing, testing, and commentary.
The short-term implications for the gaming industry include enhancing the productivity of game testers and enhancing quality assurance processes. 
One possible negative impact of such advancement is the facilitation of in-game cheating. As AI models becomes more adept at understanding game contents, there is a risk that they could be used to create sophisticated cheating tools.

\textbf{Biases and mistakes in dataset generation:}
In this study, we use existing models to annotate our dataset.
While this approach aligns with previous studies~\cite{liu2023llava, zhu2024llava} and follows the same principles as the teacher-student learning paradigm~\cite{hinton2015distilling}, we acknowledge that our data may contain biases and mistakes introduced by these existing models.



\textbf{Quality of images:}
Images in our dataset, sourced from YouTube videos, may include low-quality frames despite filtering for high-quality videos. Motion blur in video games and sampling during scene transitions can result in some blurry or less-than-ideal images.


\textbf{Use of various models for annotation:} In the annotation process, we utilize various models depending on their availability and cost. While this process brings diversity to the labels, it is not meant to compare the different models against each other in terms of performance.

\textbf{Use of multiple-choice questions to evaluate game understanding:} We use multiple-choice questions to evaluate a model's game understanding, as it allows for a clear comparison. While this format has been extensively for benchmarks~\cite{hendrycks2020measuring, yue2024mmmu},  it might not be the best proxy for game understanding. Future work needs to focus on human evaluation or the use of LLMs as judges~\cite{zheng2024judging}.

\section{Conclusion}
We introduce a new instruction-following dataset, with 389,565 image-instruction pairs, specifically designed for video game understanding. We investigate the effectiveness of fine-tuning LMMs on different instruction-following dataset types and mixtures of them, and finally introduce \methodname, an 8B parameter model that outperforms a SOTA model, \llavaonesix, on a game-related question answering benchmark.

\bibliographystyle{splncs04}
\bibliography{work_cited}


\clearpage
\setcounter{page}{1}
\maketitlesupplementary

\newcommand{\beginsupplementary}{%
    \setcounter{table}{0}
    \renewcommand{\thetable}{A\arabic{table}}%
    
    \setcounter{figure}{0}
    \renewcommand{\thefigure}{A\arabic{figure}}%
    
    \setcounter{section}{0}
    \renewcommand{\thesection}{A\arabic{section}}
    \renewcommand{\thesubsection}{\thesection.\arabic{subsection}}
}
\beginsupplementary%
\setcounter{figure}{0}
\renewcommand{\thefigure}{A\arabic{figure}}%
\renewcommand{\theHfigure}{SuppFigureA\arabic{figure}} 

\setcounter{table}{0}
\renewcommand{\thetable}{A\arabic{table}}%
\renewcommand{\theHtable}{SuppTableA\arabic{table}} 

\setcounter{section}{0}
\renewcommand{\thesection}{A\arabic{section}}
\renewcommand{\thesubsection}{\thesection.\arabic{subsection}}
\renewcommand{\theHsection}{SuppSectionA\arabic{section}} 


\section{Additional details}
\label{app:additional_details}

\subsection{Prompts used to generated datasets }
\label{app:prompts}

\begin{custommdframed}
\begin{minipage}{\linewidth}
\captionof{figure}{Long caption generation with \gptFourV}
\label{prompt:gpr4v_long_caption}
\emph{Please provide a detailed description of the image, ensuring that no details are omitted. Describe every element you observe within the image to provide a comprehensive account of its contents. Don't be lazy and it is important to get everything well done.}
\end{minipage}
\end{custommdframed}

\begin{custommdframed}
\begin{minipage}{\linewidth}
\captionof{figure}{image-to-JSON data generation}
\label{prompt:image_to_json_gemini}
\emph{
First, provide a detailed description of the image, including every small detail possible. Next, create ten multiple-choice questions based on the content of the image. Each question should test the understanding of the image's content.
Follow this JSON format:
\{
  "description": "Full Image Description",
  "short\_description": "Short Image Description",
  "dialogue": ["Any visible dialogue text as a json list"],
  "on\_screen\_subtitle": "any subtitle on the image or n/a",
  "minimap\_details": "Information from the minimap",
  "inventory\_display": "Information about the player's inventory",
  "score\_or\_progress": "Details about scores or progress indicators",
  "NPC\_status": "Information about NPCs",
  "event\_indicators": "Indicators of any special events",
  "interaction\_prompts": "Visible prompts for player interactions",
  "game\_mode": "Current game mode or context",
  "HUD\_description": "description of the game HUD or n/a if there is no HUD",
  "on\_screen\_watermark": "any watermark on the image or n/a",
  "summary\_of\_ui\_values": "summary of the UI values as json or empty json if there is no UI",
  "scene\_description": "A high-level overview of the entire scene",
  "character\_list": [
    \{
      "name": "Character Name",
      "appearance": "Description of appearance",
      "clothing": "Description of clothing",
      "facial\_expression": "Description of facial expression"
    \}
  ],
  "object\_list": ["Object 1", "Object 2", ...],
  "texture\_details": "a json list of object name and texture patterns that they have",
  "lighting\_details": "Specific information about the light sources and shadows in the scene",
  "color\_palette": ["hexadecimal color code", "hexadecimal color code", ...],
  "weather\_conditions": "Description of any weather effects present, or say cannot be determined",
  "environmental\_effects": "Description of any environmental effects like fog, rain, fire, etc.",
  "animation\_states": "Descriptions of any static poses or actions implied by character positions",
  "error\_log": "Any noticeable glitches or anomalies in the image",
  "glitches": "any glitch or buggy aspect of the image or none if there is nothing",
  "player\_status": \{
    "health": "Player's health value",
    "equipment": "Player's equipment details",
    "other\_status": "Other status indicators"
  \}
\}
}
\end{minipage}
\end{custommdframed}

\begin{custommdframed}
\begin{minipage}{\linewidth}
\captionof{figure}{LLama-3-based data generation}
\label{prompt:llama3}
\emph{Using the image description provided below, create 10 questions and their corresponding answers that pertain exclusively to the details given in the description. Format your response using JSON. \newline
\textbf{Image description}:  \textless image description here  \textgreater \newline
Ensure your questions are relevant and directly related to the image description. For example, do not ask about elements not explicitly mentioned in the description.
}
\end{minipage}
\end{custommdframed}

\begin{custommdframed}
\begin{minipage}{\linewidth}
\captionof{figure}{\gptFourO-based data generation}
\label{prompt:gpt4o}
\emph{First, provide a detailed description of the image, including every small detail possible. Next, create 10 questions answers based on the content of the image. Each question should test the understanding of the image's content.
}
\end{minipage}
\end{custommdframed}

\begin{table*}[ht]
    \centering
    \caption{Description of entries in the JSON structure }
        \begin{tabular}{ll}
            \toprule
            \textbf{Key} & \textbf{Description} \\ \midrule
            Description & Detailed description of the image \\ 
            Short description & Concise description of the image \\ 
            Dialogue & A (JSON) list containing any visible dialogue text \\ 
            On-Screen subtitle & Subtitles displayed on the image \\ 
            Inventory display & Details of the player's inventory visible on the image \\ 
            HUD description & Description of the game's Head-Up Display (HUD) \\ 
            Scene description & High-level overview of the entire scene \\ 
            NPC status & High level information about non-playable characters (NPCs) \\ 
            Character list & List of characters, including their appearances, clothing, and facial expressions \\ 
            Animation states & Descriptions of static poses or actions suggested by character positions \\ 
            Object list & A (JSON) list containing all the visible objects in the scene \\ 
            Texture details & A (JSON) list detailing object names and their texture patterns \\ 
            Lighting details & Specific information about the light sources and shadows in the scene \\ 
            Weather conditions & Description of any weather effects present, or state if they cannot be determined \\ 
            Environmental effects & Description of environmental effects such as fog, rain, or fire \\ 
            Player status & Player's health, equipment details, and other status indicators \\ \bottomrule
        \end{tabular}
    \label{supptab:json_keys}
\end{table*}

\begin{figure}[ht]
    \centering
    \includegraphics[width=\linewidth]{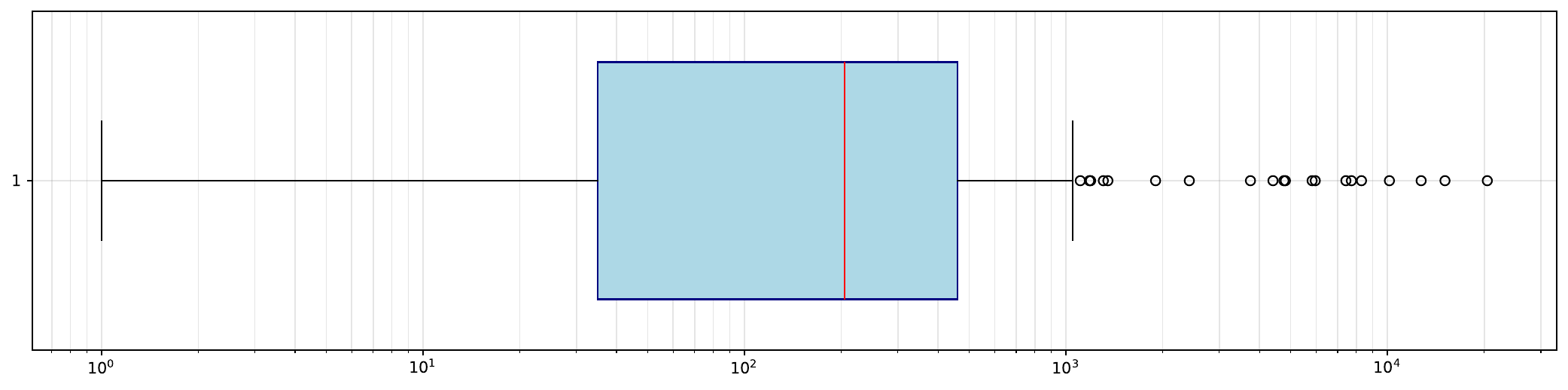}
\caption{Image distribution across games, with a median of 205 unique images per game.}
\label{suppfig:histo_games}
\end{figure}

\clearpage
\section{Additional results }
\label{app:additional_results}

In this section, we provide complementary results for the experiments conducted in the main text.

\begin{figure}[ht]
    \centering
    \includegraphics[width=\linewidth]{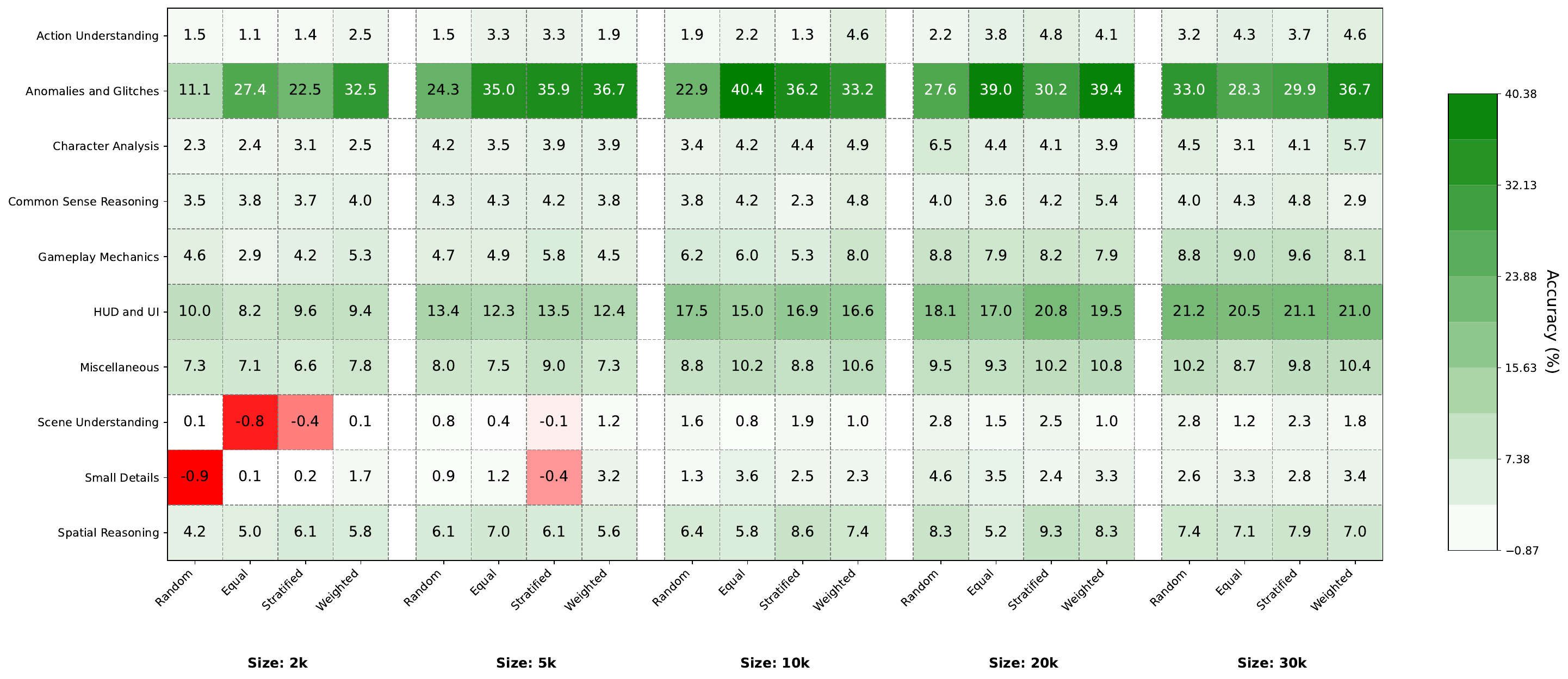}
\caption{Breakdown of improvement (percentage points) compared to the basemodel \bunny, after fine-tuning it on different mixture strategies by different question categories. As we increase the size of the dataset, all categories show improvement, with \emph{Anomalies and Glitches} showing the greatest gains.}
\label{suppfig:results_category}
\end{figure}

\begin{table}[htbp]
\centering
\small
\caption{Comparison of \methodname and \llavaonesix}
\label{apptab:vgb_vs_llava}
\begin{tabular}{l@{\hspace{7em}}r@{\hspace{1em}}l@{\hspace{3.5em}}r}
\toprule
\textbf{Category} & \multicolumn{2}{l}{\textbf{\methodname}} & \multicolumn{1}{c}{\textbf{\llavaonesix}} \\
\midrule
Action Understanding & 84.6 & \textcolor{green}{$\uparrow$}(+2.9) & 81.7 \\
Anomalies and Glitches & 82.1 & \textcolor{green}{$\uparrow$}(+16.6) & 65.5 \\
Character Analysis & 84.9 & \textcolor{green}{$\uparrow$}(+0.6) & 84.3 \\
Common Sense Reasoning & 89.1 & \textcolor{red}{$\downarrow$}(-2.8) & 91.9 \\
Gameplay Mechanics & 80.6 & \textcolor{red}{$\downarrow$}(-5.8) & 86.4 \\
HUD and UI & 84.1 & \textcolor{green}{$\uparrow$}(+3.0) & 81.1 \\
Miscellaneous & 86.2 & \textcolor{green}{$\uparrow$}(+1.7) & 84.5 \\
Scene Understanding & 92.4 & \textcolor{green}{$\uparrow$}(+0.2) & 92.2 \\
Small Details & 80.1 & \textcolor{red}{$\downarrow$}(-0.2) & 80.3 \\
Spatial Reasoning & 78.3 & \textcolor{green}{$\uparrow$}(+1.0) & 77.3 \\
\bottomrule
\end{tabular}
\end{table}

\clearpage
\section{Qualitative samples }
\label{app:bugsbunny_qual_samples}

In this section, we provide qualitative results from \methodname.

\begin{figure}[htpb]
\begin{ebox}{Qualitative results where \methodnameWhite provides a detailed description of the image.}
\centering
\includegraphics[width=\textwidth]{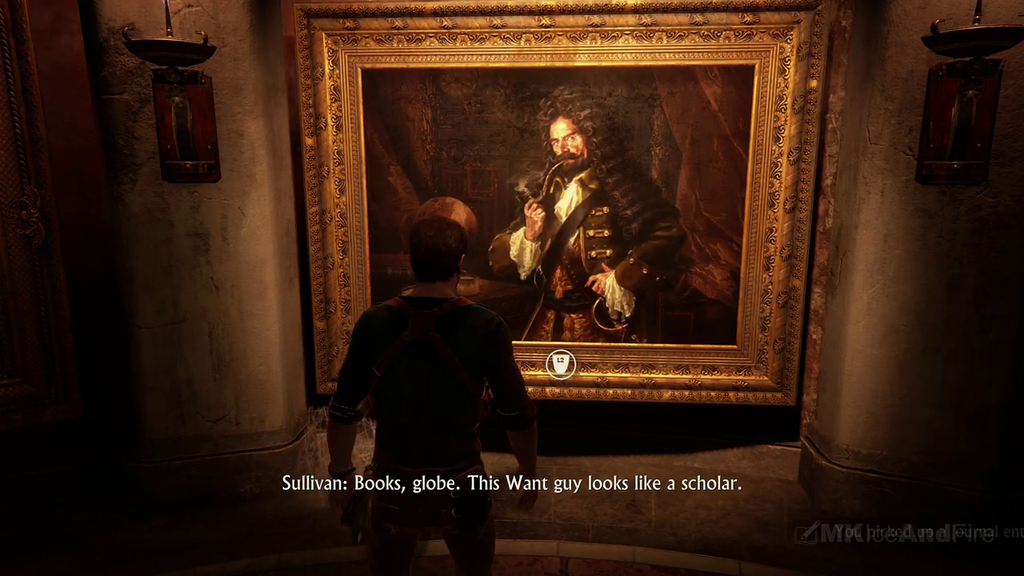}
\raggedright
\raggedright
\small
\textcolor{darkblue}{\textbf{Question:}} \textbf{Briefly describe the image.} \\
\textbf{\methodname:} The image is a screenshot from a video game, showing a character standing in front of a large, ornate painting. The painting depicts a man with long hair and a beard, wearing a period outfit with a high collar and a sash. The character is facing the painting, with their back to the viewer, and appears to be observing it. The room has a dimly lit, old-fashioned ambiance, with a sconce on the wall and a large, framed painting on the wall behind the character.
\end{ebox}
\caption{\methodname provides a detailed description of the image}
\label{suppfig:qual_detailed_okay_1}
\end{figure}

\begin{figure}[t]
\begin{ebox}{Qualitative results where \methodnameWhite provides a detailed description of the image.}
\centering
\includegraphics[width=\textwidth]{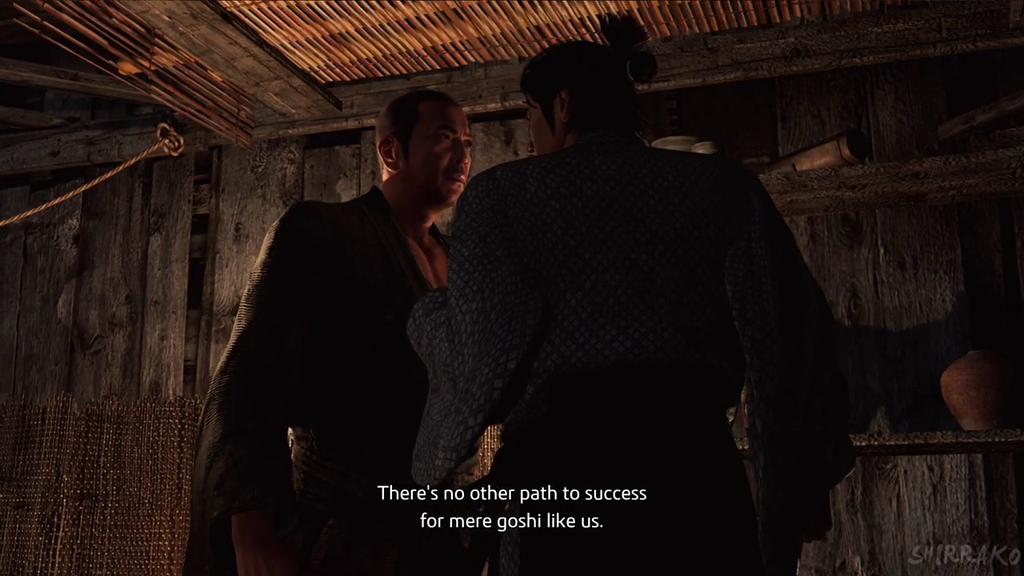}
\raggedright
\raggedright
\small
\textcolor{darkblue}{\textbf{Question:}} \textbf{Briefly describe the image.} \\
\textbf{\methodname:}  The image is a still from a video game, depicting two characters in a dimly lit wooden structure. The character on the left is a man with a stern expression, wearing a dark robe with a patterned design. He is facing the other character, who is partially obscured and has their back to the viewer. The second character appears to be wearing a similar robe. The scene is set in a rustic environment with wooden walls and a thatched roof. 
\end{ebox}
\caption{\methodname provides a detailed description of the image}
\label{suppfig:qual_detailed_okay_2}
\end{figure}

\clearpage

\clearpage
\begin{figure}[t]
\begin{ebox}{Qualitative results where \methodnameWhite provides a correct answer with minor inaccuracies}
\centering
\includegraphics[width=\textwidth]{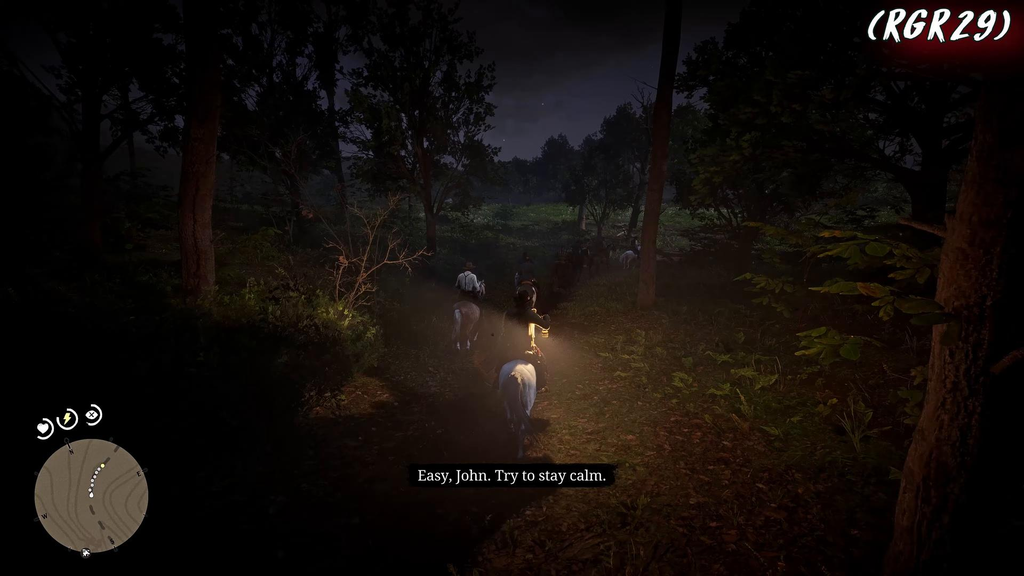}
\raggedright
\raggedright
\small
\textcolor{darkblue}{\textbf{Question:}} \textbf{Describe the characters in the image} \\
\textbf{\methodname:} The image shows two characters on horseback. The character in the foreground is holding a lantern, and the character in the background \todo{\emph{is following them}}. Both characters are wearing hats.
\end{ebox}
\caption{\methodname provides a detailed description of the image with minor inaccuracies. Details that are incorrect are highlighted and emphasized in \todo{\emph{red}}.}
\label{suppfig:qual_detailed_incorrect_1}
\end{figure}

\clearpage
\begin{figure}[t]
\begin{ebox}{Qualitative results where \methodnameWhite provides a correct answer with some inaccuracies}
\centering
\includegraphics[width=\textwidth]{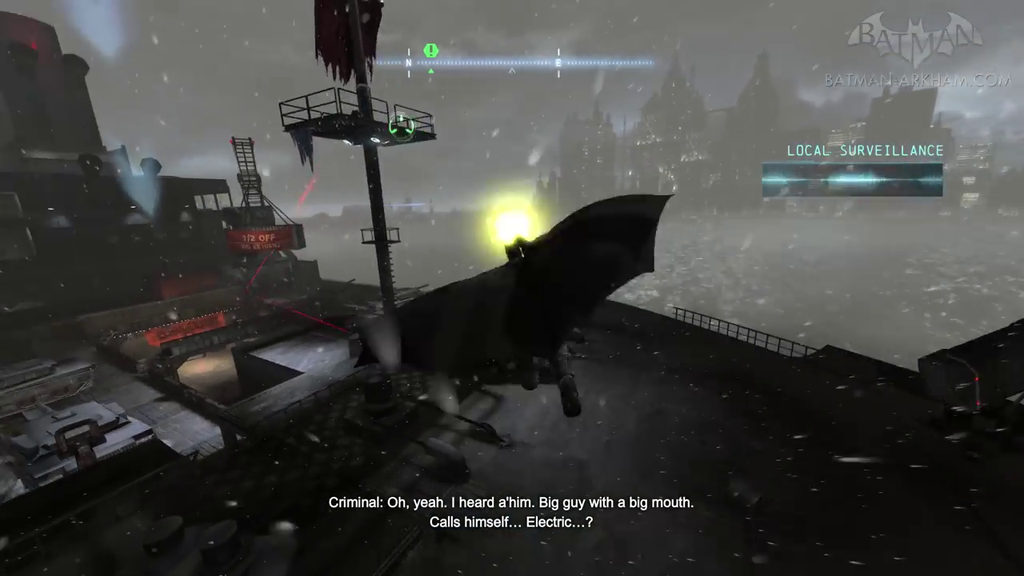}
\raggedright
\raggedright
\small
\textcolor{darkblue}{\textbf{Question:}} \textbf{Describe the characters in the image} \\
\textbf{\methodname:} The image shows Batman, who is in the center of the frame, flying through the air. He is wearing his iconic black suit\todo{\emph{ with a yellow utility belt}}. \todo{\emph{Below him, there is a character named Penguin, who is wearing a tuxedo and a bowler hat.}}
\end{ebox}
\caption{\methodname hallucinates extra characters and a Batman utility belt in the image, which are not actually present. Details that are incorrect are highlighted and emphasized in \todo{\emph{red}}.}
\label{suppfig:qual_detailed_incorrect_2}
\end{figure}

\clearpage
\begin{figure}[t]
\begin{ebox}{Qualitative results where \methodnameWhite provide a \todo{wrong} answer for a query}
\centering
\includegraphics[width=\textwidth]{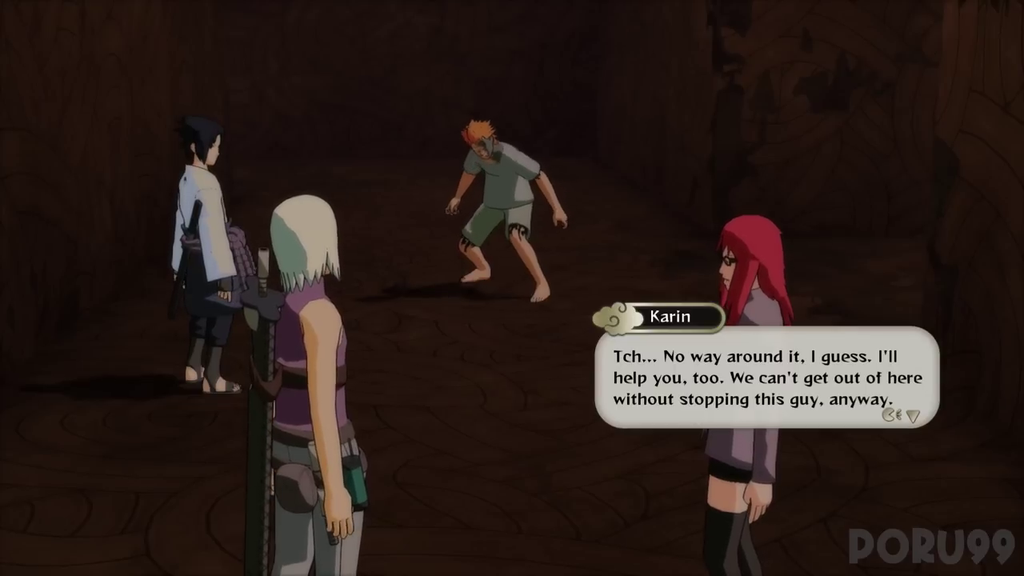}
\raggedright
\raggedright
\small
\textcolor{darkblue}{\textbf{Question:}} \textbf{What is written on the image?} \\
\textbf{\methodname:} The text on the image reads: `Tch... Tch... Tch... You way around it, I guess. Tch... Tch... Tch... I'll tell you, without stepping this guy away...'
\end{ebox}
\caption{A sample showing that \methodname struggles to convert the image to text.}
\label{suppfig:wrong_1}
\end{figure}

\begin{figure}[t]
\begin{ebox}{Qualitative results where \methodnameWhite provide a \todo{wrong} answer for a query}
\centering
\includegraphics[width=\textwidth]{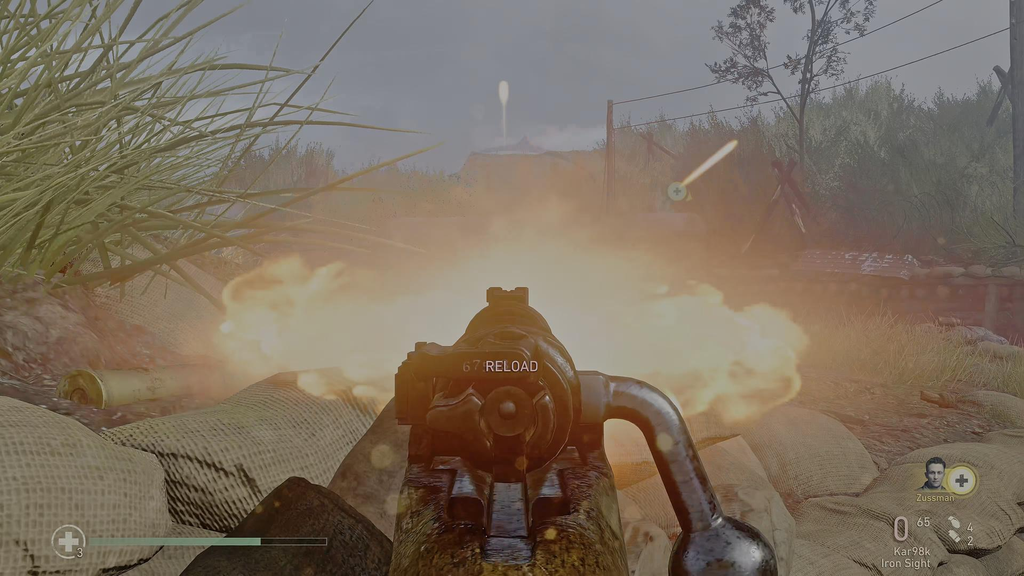}
\raggedright
\raggedright
\small
\textcolor{darkblue}{\textbf{Question:}} \textbf{What is written on the left?} \\
\textbf{\methodname:} The word 'RELOAD' is written on the left side of the gun.
\end{ebox}
\caption{A sample showing that \methodname fails to provide a correct answer. While the word \textit{RELOAD} is written in the center of the image, the query asks, \textit{What is written on the left?} which should be answered with \textit{nothing}.}
\label{suppfig:wrong_2}
\end{figure}

\clearpage

\section{Sample image-to-JSON conversion}
\label{appsec:img2json}

\begin{tcolorbox}[
    breakable,
    colback=white,
    colframe=black,
    title=Sample image-to-JSON conversion done by \methodnameWhite
]
\begin{center}
    \includegraphics[width=0.9\textwidth]{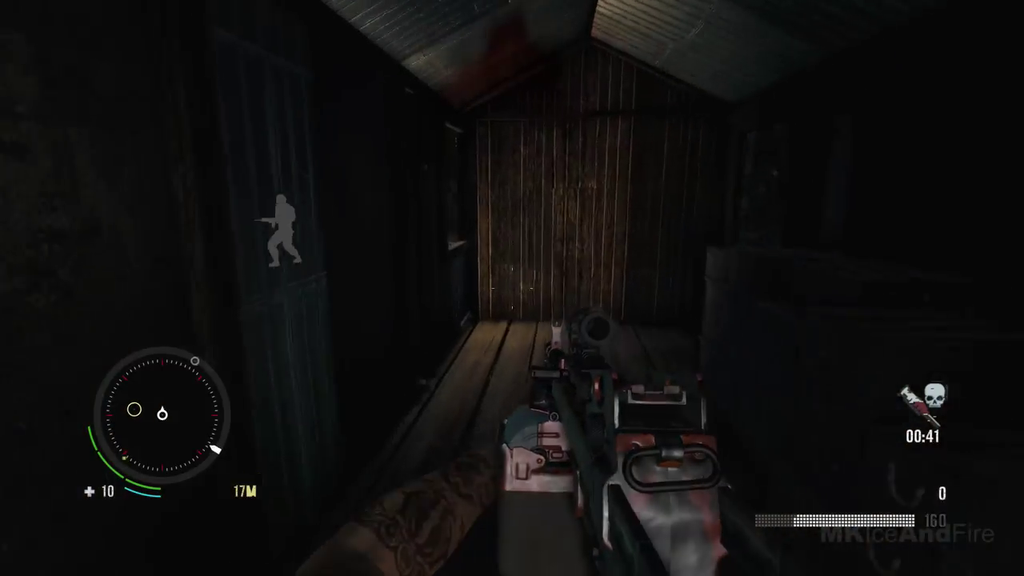}
\end{center}

\vspace{1em}

\begin{minted}[
    breaklines,
    breakanywhere,
    fontsize=\tiny,
    numbersep=5pt,
    frame=lines,
    framesep=2mm,
    linenos
]{json}
{
  "description": "The image shows a first-person perspective of a video game, likely a survival horror title. The player is holding a large, futuristic-looking rifle with a circular scope. The rifle has a white and black camouflage pattern. The player's left hand is visible, adorned with a tattoo of a skull and what appears to be a snake. The player is standing in a dark, dilapidated wooden structure, possibly a barn or shed. The walls are made of weathered wooden planks, and the floor is made of wooden planks as well. There is a doorway on the left side of the image, leading to a darker area. A small, white silhouette of a person holding a rifle is visible on the left side of the screen, indicating the player's position on a minimap. The minimap shows the player's location as a white dot within a circular area. The player's health bar is visible at the bottom of the screen, represented by a green bar with a white outline. The health bar is nearly full. The player's ammunition count is displayed in the bottom right corner of the screen, showing '0' bullets remaining in the current magazine and '41' bullets in reserve. A small icon of a skull with a knife is visible next to the ammunition count, possibly indicating a special ability or weapon. The overall atmosphere of the image is dark, gritty, and suspenseful, typical of the survival horror genre.",
  "dialogue": [],
  "on_screen_subtitle": "n/a",
  "minimap_details": "The minimap shows the player's location as a white dot within a circular area. The minimap is located in the bottom left corner of the screen.",
  "inventory_display": "The image does not show the player's full inventory.",
  "score_or_progress": "The image does not display any score or progress indicators.",
  "NPC_status": "The image does not provide any information about NPCs.",
  "event_indicators": "The image does not display any specific event indicators.",
  "interaction_prompts": "The image does not show any interaction prompts.",
  "game_mode": "The image suggests the game is in a single-player campaign mode, based on the presence of a minimap and health bar.",
  "HUD_description": "The HUD consists of a circular minimap in the bottom left corner, a circular health bar in the bottom left corner, and an ammunition counter in the bottom right corner. The ammunition counter displays the current magazine capacity, the remaining bullets in the magazine, and the total reserve ammunition.",
  "on_screen_watermark": "MKiceAndFire",
  "summary_of_ui_values": {
    "health": "Nearly full",
    "ammunition": "0/41"
  },
  "scene_description": "The scene takes place inside a dilapidated wooden structure, likely a barn or shed. The structure is dark and appears to be abandoned. The walls and floor are made of weathered wooden planks. There is a doorway on the left side of the image, leading to a darker area.",
  "character_list": [
    {
      "name": "Player",
      "appearance": "Cannot be fully determined, but the player's left hand is visible, showing a tattoo of a skull and a snake.",
      "clothing": "Cannot be fully determined",
      "facial_expression": "Cannot be determined"
    }
  ],
  "object_list": [
    "Rifle",
    "Scope",
    "Wooden planks (walls and floor)",
    "Doorway"
  ],
  "texture_details": [
    {
      "object": "Rifle",
      "texture": "White and black camouflage pattern"
    },
    {
      "object": "Wooden planks",
      "texture": "Weathered, with visible grain and knots"
    }
  ],
  "lighting_details": "The scene is dimly lit, with the primary light source appearing to be coming from the doorway on the left. This creates a sense of depth and shadows within the structure.",
  "color_palette": [
    "#181818",
    "#303030",
    "#505050",
    "#FFFFFF"
  ],
  "weather_conditions": "Cannot be determined",
  "environmental_effects": "None",
  "animation_states": "The player is holding the rifle in a ready position, suggesting a state of alertness and preparedness for combat.",
  "error_log": "None",
  "glitches": "None",
  "player_status": {
    "health": "Nearly full",
    "equipment": "The player is equipped with a large, futuristic-looking rifle with a circular scope.",
    "other_status": "The player has 0 bullets remaining in the current magazine and 41 bullets in reserve."
  }
}
\end{minted}
\end{tcolorbox}

\begin{tcolorbox}[
    breakable,
    colback=white,
    colframe=black,
    title=Sample image-to-JSON conversion done by \methodnameWhite
]
\begin{center}
    \includegraphics[width=0.9\textwidth]{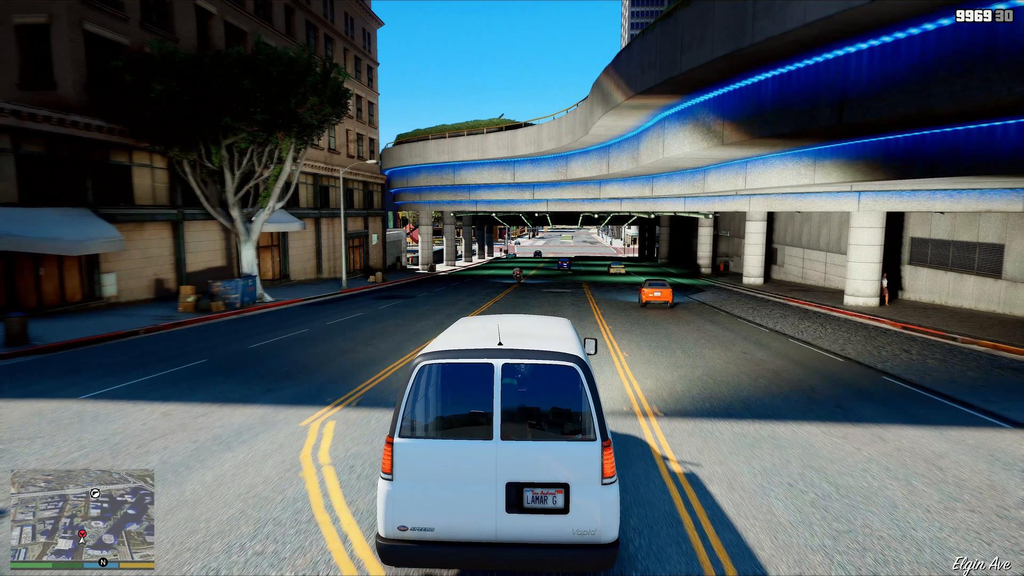}
\end{center}

\vspace{1em}

\begin{minted}[
    breaklines,
    breakanywhere,
    fontsize=\tiny,
    numbersep=5pt,
    frame=lines,
    framesep=2mm,
    linenos
]{json}
{
  "description": "The image shows a third-person perspective of a video game, likely Grand Theft Auto V, based on the HUD elements and graphics style. The player is driving a white van on a multi-lane road. The van is positioned slightly to the right of the center of the road, heading towards the viewer. The road is a typical city street with sidewalks, buildings, and other vehicles. The time of day appears to be midday with clear skies. The HUD elements visible include a minimap in the bottom left corner, a money counter in the top right corner, and a small portion of the player's health bar in the bottom right corner. The minimap shows the player's location on a simplified map of the surrounding area. The money counter displays the amount of in-game currency the player has, which is 9969. The health bar is partially obscured, but it appears to be mostly full.",
  "dialogue": [],
  "on_screen_subtitle": "n/a",
  "minimap_details": "The minimap shows the player's location on a simplified map of the surrounding area. The player is represented by a white arrow pointing upwards, indicating their direction. The map shows roads, buildings, and a body of water, likely a river or lake.",
  "inventory_display": "Not visible in the image.",
  "score_or_progress": "The player's current in-game money is displayed as 9969 in the top right corner.",
  "NPC_status": "Not applicable, as there are no visible NPCs in the image.",
  "event_indicators": "None visible in the image.",
  "interaction_prompts": "None visible in the image.",
  "game_mode": "Cannot be determined from the image.",
  "HUD_description": "The HUD consists of a minimap in the bottom left corner, a money counter in the top right corner, and a small portion of the player's health bar in the bottom right corner.",
  "on_screen_watermark": "The watermark \"Elegy Ave\" is visible in the bottom right corner.",
  "summary_of_ui_values": {
    "money": "9969"
  },
  "scene_description": "The scene depicts a typical urban environment with buildings, roads, and vehicles. The player is driving a van on a multi-lane road, heading towards a large, modern-looking building with a blue light display.",
  "character_list": [],
  "object_list": [
    "Van",
    "Buildings",
    "Road",
    "Sidewalk",
    "Trees",
    "Vehicles",
    "Streetlights",
    "Blue Light Display"
  ],
  "texture_details": [
    {
      "object": "Buildings",
      "texture": "Brick, concrete, glass"
    },
    {
      "object": "Road",
      "texture": "Asphalt"
    },
    {
      "object": "Sidewalk",
      "texture": "Concrete"
    }
  ],
  "lighting_details": "The scene is brightly lit, suggesting midday sun. The lighting casts soft shadows from the buildings and trees onto the road.",
  "color_palette": [
    "#87CEEB",
    "#FFFFFF",
    "#A9A9A9",
    "#696969",
    "#808080"
  ],
  "weather_conditions": "Clear and sunny.",
  "environmental_effects": "None visible in the image.",
  "animation_states": "The van's wheels suggest it is in motion, and the other vehicles on the road imply traffic.",
  "error_log": "None visible in the image.",
  "glitches": "None visible in the image.",
  "player_status": {
    "health": "Partially obscured, but appears to be mostly full.",
    "equipment": "Not visible in the image.",
    "other_status": "The player has 9969 in-game currency."
  }
}
\end{minted}
\end{tcolorbox}

\begin{tcolorbox}[
    breakable,
    colback=white,
    colframe=black,
    title=Sample image-to-JSON conversion done by \methodnameWhite
]
\begin{center}
    \includegraphics[width=0.9\textwidth]{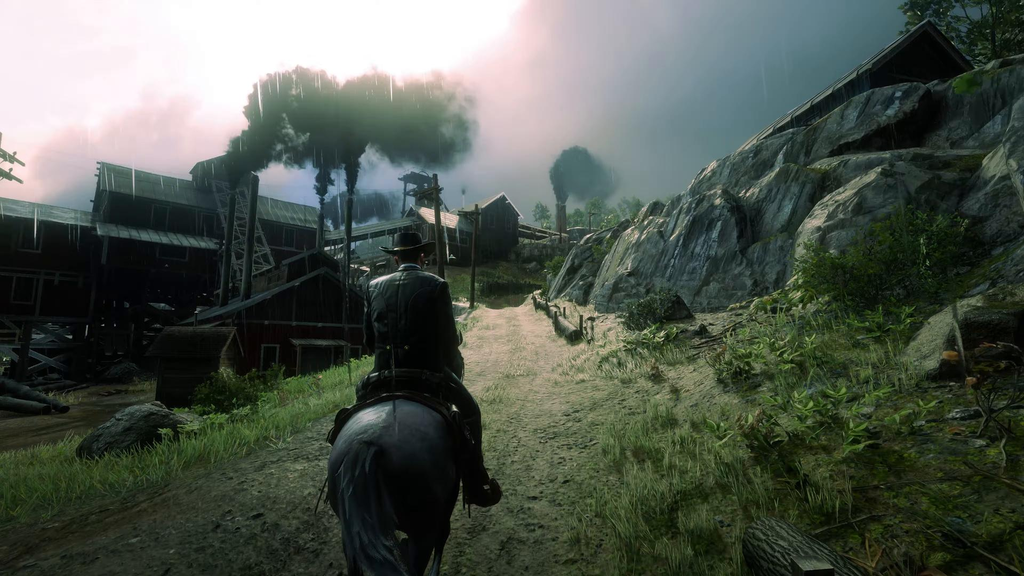}
\end{center}

\vspace{1em}

\begin{minted}[
    breaklines,
    breakanywhere,
    fontsize=\tiny,
    numbersep=5pt,
    frame=lines,
    framesep=2mm,
    linenos
]{json}
{
  "description": "The image captures a scene from the video game Red Dead Redemption 2, showcasing the protagonist, Arthur Morgan, riding his horse, Black Jack, along a dirt path. The path leads towards a small, industrial settlement built on a cliffside. The settlement features wooden buildings with corrugated metal roofs, characteristic of the late 19th-century American frontier. A plume of black smoke rises from a chimney in the background, suggesting industrial activity. The surrounding environment is a mix of rocky terrain and lush vegetation, with a dense forest visible in the distance. The sky is overcast, with a hazy, diffused light suggesting either early morning or late afternoon. The overall atmosphere is one of quiet anticipation, as if Arthur is about to enter a new chapter in his journey.",
  "dialogue": [],
  "on_screen_subtitle": "n/a",
  "minimap_details": "n/a",
  "inventory_display": "n/a",
  "score_or_progress": "n/a",
  "NPC_status": "n/a",
  "event_indicators": "n/a",
  "interaction_prompts": "n/a",
  "game_mode": "n/a",
  "HUD_description": "n/a",
  "on_screen_watermark": "n/a",
  "summary_of_ui_values": {},
  "scene_description": "Arthur Morgan riding his horse Black Jack towards a small industrial settlement on a cliffside.",
  "character_list": [
    {
      "name": "Arthur Morgan",
      "appearance": "A rugged man with a beard and mustache, wearing a dark hat and a dark coat.",
      "clothing": "Dark hat, dark coat, dark pants, and boots.",
      "facial_expression": "Cannot be determined as his back is to the camera."
    }
  ],
  "object_list": [
    "Horse",
    "Dirt path",
    "Wooden buildings",
    "Corrugated metal roofs",
    "Smoke plume",
    "Chimney",
    "Forest",
    "Rocks",
    "Vegetation",
    "Trees"
  ],
  "texture_details": [
    {
      "object": "Horse",
      "texture": "Short, dark brown fur"
    },
    {
      "object": "Dirt path",
      "texture": "Uneven, dusty, with patches of grass"
    },
    {
      "object": "Wooden buildings",
      "texture": "Weathered wood planks"
    },
    {
      "object": "Corrugated metal roofs",
      "texture": "Rusted and dented metal sheets"
    },
    {
      "object": "Rocks",
      "texture": "Rough, jagged, with moss and lichen"
    },
    {
      "object": "Vegetation",
      "texture": "A mix of grass, shrubs, and trees"
    }
  ],
  "lighting_details": "The scene is lit by diffused sunlight, suggesting an overcast sky. The light is soft and even, with no strong shadows.",
  "color_palette": [
    "#A0A0A0",
    "#808080",
    "#606060",
    "#404040",
    "#202020",
    "#008000",
    "#006400",
    "#004000"
  ],
  "weather_conditions": "Overcast",
  "environmental_effects": "Smoke plume rising from a chimney",
  "animation_states": "Arthur Morgan is sitting upright on his horse, suggesting a calm and controlled ride.",
  "error_log": "None",
  "glitches": "None",
  "player_status": {
    "health": "n/a",
    "equipment": "n/a",
    "other_status": "n/a"
  }
}
\end{minted}
\end{tcolorbox}

\begin{tcolorbox}[
    breakable,
    colback=white,
    colframe=black,
    title=Sample image-to-JSON conversion done by \methodnameWhite
]
\begin{center}
    \includegraphics[width=0.9\textwidth]{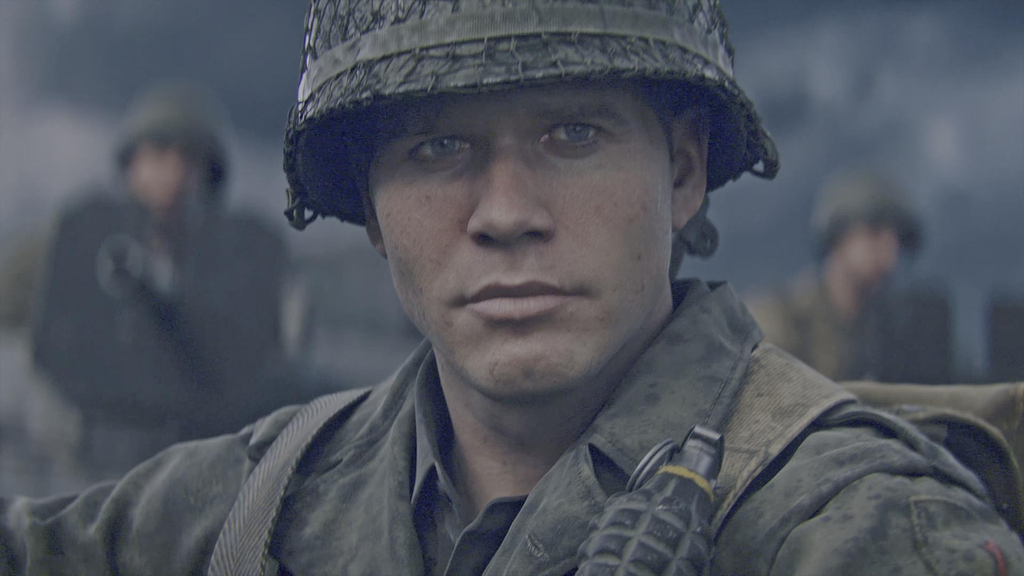}
\end{center}

\vspace{1em}

\begin{minted}[
    breaklines,
    breakanywhere,
    fontsize=\tiny,
    numbersep=5pt,
    frame=lines,
    framesep=2mm,
    linenos
]{json}
{
  "description": "The image showcases a close-up of a man in a World War II-era military uniform. He is positioned slightly to the right of the center, his gaze directed slightly upwards and to the left. His expression is serious, bordering on somber. He has short, dark hair and a prominent nose. The man's uniform is a muted green, with a noticeable dirt smudge on his left cheek. A dark green helmet with a netting pattern covers his head, partially obscuring his hair. The helmet's chin strap is visible, fastened with a metal buckle. A thick, dark green strap crosses his chest, likely part of a backpack or equipment harness. The background is a blurred depiction of a battlefield, with other soldiers in similar uniforms visible in the distance. The overall color palette is muted and desaturated, reflecting the grim atmosphere of war.",
  "dialogue": [],
  "on_screen_subtitle": "n/a",
  "minimap_details": "n/a",
  "inventory_display": "n/a",
  "score_or_progress": "n/a",
  "NPC_status": "n/a",
  "event_indicators": "n/a",
  "interaction_prompts": "n/a",
  "game_mode": "n/a",
  "HUD_description": "n/a",
  "on_screen_watermark": "n/a",
  "summary_of_ui_values": {},
  "scene_description": "The scene depicts a soldier in the midst of a battlefield, likely during World War II, given the uniforms and equipment.",
  "character_list": [
    {
      "name": "Unknown",
      "appearance": "The man appears to be in his late twenties to early thirties. He has a rugged appearance with short, dark hair and a prominent nose.",
      "clothing": "He is wearing a World War II-era military uniform in a muted green color. A dark green helmet with a netting pattern covers his head, and a thick, dark green strap crosses his chest, likely part of a backpack or equipment harness.",
      "facial_expression": "His expression is serious and slightly somber, suggesting he is focused on the situation at hand."
    }
  ],
  "object_list": [
    "Military uniform",
    "Helmet",
    "Backpack strap",
    "Backpack (partially visible)",
    "Soldiers (in the background)"
  ],
  "texture_details": [
    {
      "object": "Helmet",
      "texture": "Netting pattern"
    },
    {
      "object": "Uniform",
      "texture": "Camouflage pattern"
    }
  ],
  "lighting_details": "The lighting is soft and diffused, likely due to overcast skies or fog. It casts subtle shadows on the man's face, accentuating his features.",
  "color_palette": [
    "#5C6B5A",
    "#3D4A3B",
    "#7A8A76",
    "#B2B4A9"
  ],
  "weather_conditions": "The weather appears to be overcast or foggy, as evidenced by the soft lighting and lack of harsh shadows.",
  "environmental_effects": "The background suggests a battlefield, with smoke or dust particles visible in the air.",
  "animation_states": "The man appears to be in a still pose, likely captured in a moment of pause or reflection during the chaos of battle.",
  "error_log": "None",
  "glitches": "None",
  "player_status": {
    "health": "n/a",
    "equipment": "n/a",
    "other_status": "n/a"
  }
}
\end{minted}
\end{tcolorbox}

\begin{tcolorbox}[
    breakable,
    colback=white,
    colframe=black,
    title=Sample image-to-JSON conversion done by \methodnameWhite
]
\begin{center}
    \includegraphics[width=0.9\textwidth]{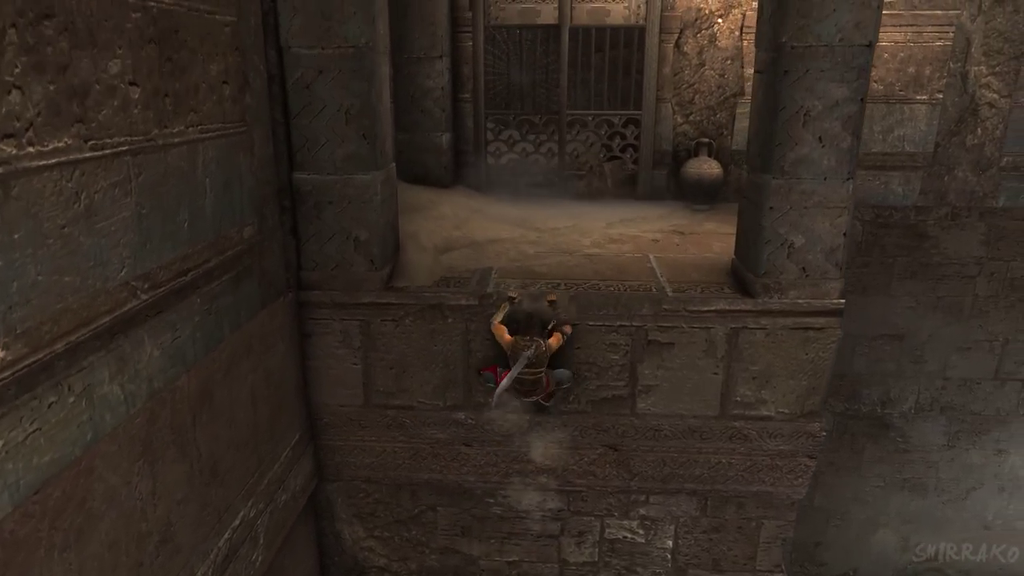}
\end{center}

\vspace{1em}

\begin{minted}[
    breaklines,
    breakanywhere,
    fontsize=\tiny,
    numbersep=5pt,
    frame=lines,
    framesep=2mm,
    linenos
]{json}
{
  "description": "The image shows a male character in a video game, seemingly in the middle of a climbing or jumping action. He is positioned on a stone platform within a dimly lit, ancient-looking structure. The platform is part of a larger architectural complex, with visible stone walls and a barred gate in the background. The character is dressed in a sleeveless top and pants, suggesting a warrior or adventurer archetype. He is equipped with a sword sheathed on his back and appears to be in motion, with his body angled towards the platform's edge. The overall atmosphere is one of adventure and exploration, with the character's pose hinting at a moment of action or transition within the game.",
  "dialogue": [],
  "on_screen_subtitle": "n/a",
  "minimap_details": "Cannot be determined",
  "inventory_display": "Not visible",
  "score_or_progress": "Not visible",
  "NPC_status": "No NPCs present",
  "event_indicators": "None visible",
  "interaction_prompts": "None visible",
  "game_mode": "Cannot be determined",
  "HUD_description": "n/a",
  "on_screen_watermark": "SHIRRAKO in the bottom right corner",
  "summary_of_ui_values": {},
  "scene_description": "The scene is set within a large, ancient-looking structure, possibly a temple or palace. The architecture suggests a Middle Eastern or Mediterranean influence. The environment is dimly lit, with the primary light source appearing to be natural light filtering in from an unseen opening above.",
  "character_list": [
    {
      "name": "Unknown",
      "appearance": "Male, athletic build",
      "clothing": "Sleeveless top, pants, possibly leather or fabric",
      "facial_expression": "Cannot be determined"
    }
  ],
  "object_list": [
    "Stone platform",
    "Stone walls",
    "Barred gate",
    "Sword",
    "Pottery jug"
  ],
  "texture_details": [
    {
      "object": "Stone walls",
      "texture": "Rough, uneven stone with visible cracks and weathering"
    },
    {
      "object": "Stone platform",
      "texture": "Similar to the walls, but with a slightly smoother appearance"
    }
  ],
  "lighting_details": "The scene is dimly lit, with the primary light source appearing to be natural light coming from an unseen opening above. This creates a sense of depth and highlights the textures of the stone surfaces.",
  "color_palette": [
    "#5C4A3A",
    "#3B322A",
    "#7A6D5B",
    "#A1887F"
  ],
  "weather_conditions": "Cannot be determined",
  "environmental_effects": "None visible",
  "animation_states": "The character's pose suggests a dynamic action, either in the middle of a jump or a climbing maneuver.",
  "error_log": "None visible",
  "glitches": "None",
  "player_status": {
    "health": "Not visible",
    "equipment": "Sword visible on back",
    "other_status": "Not visible"
  }
}
\end{minted}
\end{tcolorbox}



\end{document}